\documentclass[10pt,twocolumn,letterpaper]{article}

\usepackage[pagenumbers]{cvpr} 

\usepackage{graphicx}
\usepackage{amsmath}
\usepackage{amssymb}
\usepackage{booktabs}

\usepackage[numbers]{natbib}

\usepackage[utf8]{inputenc} 
\usepackage[T1]{fontenc}    
\usepackage{url}            
\usepackage{booktabs}       
\usepackage{amsfonts}       
\usepackage{nicefrac}       
\usepackage{microtype}      
\usepackage{xcolor}         
\usepackage[noend]{algpseudocode}
\usepackage{comment}

\usepackage{amsmath} 
\usepackage{bm} 
\usepackage{bbm} 
\makeatletter
\@ifundefined{proposition}{%
    
}{}
\@ifundefined{definition}{%
}{}
\@ifundefined{lemma}{%
    
}{}
\@ifundefined{corollary}{%
    \newtheorem{corollary}{Corollary}
}{}
\@ifundefined{theorem}{%
    \newtheorem{theorem}{Theorem}
}{}
\@ifundefined{corollary}{%
    
}{}
\@ifundefined{assumption}{%
    
}{}
\@ifundefined{problem}{%
    
}{}
\@ifundefined{claim}{%
    
}{}
\makeatother

\newtheorem{innercustomgeneric}{\customgenericname}
\providecommand{\customgenericname}{}
\newcommand{\newcustomtheorem}[2]{%
  \newenvironment{#1}[1]
  {%
   \renewcommand\customgenericname{#2}%
   \renewcommand\theinnercustomgeneric{##1}%
   \innercustomgeneric
  }
  {\endinnercustomgeneric}
}
\newcustomtheorem{customclaim}{Claim}

\providecommand{\eqa}				[1]		{\begin{align}#1\end{align}}
\providecommand{\eqas}			[1]		{\begin{align*}#1\end{align*}}

\providecommand{\ie}{\emph{i.e.,}~}
\providecommand{\eg}{\emph{e.g.,}~}

\providecommand\qcomment[1]{ }

\providecommand{\realnum}					{\mathbb{R}}

\renewcommand{\(}						{\left(}
\renewcommand{\)}						{\right)}
\renewcommand{\[}						{\left[}
\renewcommand{\]}						{\right]}

\providecommand{\Prob}{\mathbbm{P}}
\providecommand{\Probop}{\mathop{\Prob}}
\providecommand{\Exp}{\mathbbm{E}}
\providecommand{\Expop}{\mathop{\Exp}}

\def\bh{\hat{{b}}}

\def\fh{\hat{{f}}}

\def\yh{\hat{{y}}}

\def\Ch{\hat{{C}}}

\def\Bs{\mathcal{{B}}}
\def\Cs{\mathcal{{C}}}

\def\Gs{\mathcal{{G}}}

\def\Is{\mathcal{{I}}}

\def\Ks{\mathcal{{K}}}

\def\Ps{\mathcal{{P}}}

\def\Xs{\mathcal{{X}}}
\def\Ys{\mathcal{{Y}}}

\usepackage{mathtools} 

\newcommand{\SP}[1]{{\color{blue}[SP: #1]}}

\usepackage{xcolor}
\usepackage{subcaption}
\newcommand{\vmid}{\;\middle|\;}

\newcommand{\prp}{\text{prp}}
\newcommand{\loc}{\text{loc}}

\newcommand{\prs}{\text{prs}}
\newcommand{\dtr}{\text{det}}
\newcommand{\trac}{\text{trac}}
\newcommand{\gt}{\text{gt}}

\newcommand{\edge}{\text{edge}}
\usepackage{makecell}

\usepackage[ruled,vlined]{algorithm2e}
\usepackage{bbm}

\usepackage[pagebackref,breaklinks,colorlinks]{hyperref}

\usepackage[capitalize]{cleveref}
\crefname{section}{Sec.}{Secs.}
\Crefname{section}{Section}{Sections}
\Crefname{table}{Table}{Tables}
\crefname{table}{Tab.}{Tabs.}

\def\cvprPaperID{6690} 
\def\confName{CVPR}
\def\confYear{2022}

\usepackage{overpic}
\usepackage{enumitem} 
\usepackage{overpic} 
\usepackage{color}

\definecolor{turquoise}{cmyk}{0.65,0,0.1,0.3}
\definecolor{purple}{rgb}{0.65,0,0.65}
\definecolor{dark_green}{rgb}{0, 0.5, 0}
\definecolor{orange}{rgb}{0.8, 0.6, 0.2}
\definecolor{red}{rgb}{0.8, 0.2, 0.2}
\definecolor{darkred}{rgb}{0.6, 0.1, 0.05}
\definecolor{blueish}{rgb}{0.0, 0.3, .6}
\definecolor{light_gray}{rgb}{0.7, 0.7, .7}
\definecolor{pink}{rgb}{1, 0, 1}
\definecolor{greyblue}{rgb}{0.25, 0.25, 1}

\DeclareMathOperator*{\argmax}{arg\,max}

\usepackage{blindtext}

\renewcommand{\paragraph}[1]{\vspace{1em}\noindent\textbf{#1}.}
\begin{document}
\title{Towards PAC Multi-Object Detection and Tracking}

\author{Shuo Li\\
University of Pennsylvania\\
{\tt\small lishuo1@seas.upenn.edu}
\and 
Sangdon Park\\
Georgia Institute of Technology\\
{\tt\small sangdonp@cis.upenn.edu}
\and
Xiayan Ji\\
University of Pennsylvania\\
{\tt\small xjiae@seas.upenn.edu}
\and
Insup Lee\\
University of Pennsylvania\\
{\tt\small lee@cis.upenn.edu}
\and
Osbert Bastani\\
University of Pennsylvania\\
{\tt\small obastani@seas.upenn.edu}
}
\maketitle
\begin{abstract}
Accurately detecting and tracking multi-objects is important for safety-critical applications such as autonomous navigation. However, it remains challenging to provide guarantees on the performance of state-of-the-art techniques based on deep learning.
We consider a strategy known as conformal prediction, which predicts sets of labels instead of a single label; in the classification and regression settings, these algorithms can guarantee that the true label lies within the prediction set with high probability. Building on these ideas, we propose multi-object detection and tracking algorithms that come with probably approximately correct (PAC) guarantees. They do so by constructing both a prediction set around each object detection as well as around the set of edge transitions; given an object, the detection prediction set contains its true bounding box with high probability, and the edge prediction set contains its true transition across frames with high probability. We empirically demonstrate that our method can detect and track objects with PAC guarantees on the COCO and MOT-17 datasets.
\end{abstract}


\section{Introduction}

Multi-object detectors and trackers are key components of vision-based autonomous systems \cite{milan2016mot16,ciaparrone2020deep}. As they are increasingly deployed in safety-critical systems, it becomes important for ensuring their correctness. 
A promising strategy is uncertainty quantification \cite{smith2013uncertainty}, which has received recent attention in the context of deep learning \cite{gawlikowski2021survey}. Prediction sets quantify uncertainty in a way that can provide probably approximately correct (PAC) guarantees \cite{valiant1984theory, Park2020PAC} for individual classification or regression models. However, object detection and tracking requires more complex predictors, posing challenges to applying existing approaches.

In this paper, we propose novel compositional approaches to construct PAC prediction sets for multi-object detectors and trackers. We use PAC prediction sets as a building block \cite{Park2020PAC}, providing a PAC guarantees for the overall system by composing PAC guarantees for individual components.

For detection, we build on faster RCNN \citep{Ren2015FasterRT}, which consists of three components: the proposal, presence, and location components. Given an image, the proposal component outputs bounding boxes that likely contains objects, called \emph{proposals}. Next, the presence component outputs the likelihood of the presence of a target object class in the proposals. Then, the location component outputs the location of the target object class in the proposals. Finally, the detector returns a set of the presence and location pairs of a target class as the final detection outputs. We first apply the PAC prediction set algorithm to each component, and then compose them to provide a PAC guarantee for the overall detector.

For tracking, we build on Tracktor \citep{Bergmann_2019}, which consists of an object detection component and an object connection component. The object detection component outputs an object presence score and object location of a target class for each image, achieved by running an off-the-shelf object detector such as faster RCNN. Then, the object connection component predicts the connection, or \emph{edge}, between a pair of object detections at adjacent time frames. The edge prediction is based on their similarities, e.g., the intersection over union (IoU) between their bounding boxes. We use our PAC object detector and additional provide PAC prediction sets on the edges connecting objects across frames.

We empirically evaluate our method on the COCO data set for object detection and the MOT-17 dataset for object tracking.
For object detection, we demonstrate that the component-wise  PAC prediction sets satisfy the PAC guarantee individually; moreover, the composed detection prediction set also satisfies the PAC guarantee from the individual guarantee.
For multi-object tracking, we demonstrate that our PAC predicion sets outperform a na\"{i}ve baseline that simply uses the top 1-5 bounding boxes with the highest scores. In particular, our PAC set always meets the desired error rate guarantee while maintaining a lower false-positive rate than the baselines that satisfy this guarantee.


\section{Related Work}

\textbf{Object detection.}
One-stage detectors typically use a fully convolutional design, with the network's outputs being classification probabilities and box offsets for each spatial location, for example YOLO \cite{Redmon2016YouOL, redmon2013darknet,redmon2017yolo9000, bochkovskiy2020yolov4}, RetinaNet~\cite{lin2017focal}, Single Shot Detector~\cite{Liu2016SSDSS}, and Swin Transformer~\cite{liu2021swin}.

We use a two-stage detector since the component-wise design makes it easier to incorporate prediction sets.
A typical two-stage detector first uses an region proposal network (RPN) to propose regions that likely contain objects, and then feeds these to a region convolutional network (R-CNN) to obtain classification scores and spatial offsets. \cite{Girshick2015FastR} introduces Fast R-CNN to optimize RCNN-training procedure. Faster R-CNN \cite{Ren2015FasterRT} reduced computation speed by integrating the region proposal algorithm into the CNN model, and was further extended by Mask R-CNN \cite{he2017mask}.
More recent two-stage detectors compute object proposals first and use them to extract features and detect objects \cite{chen2019hybrid,cai2018cascade,qiao2021detectors}.
Existing object detectors do not provide PAC guarantees on detection results, whereas our goal is to provide such guarantees. Intuitively, we guarantee that all objects are covered with high probability (though there might be spurious detections).

\textbf{Object tracking.}
Most recent multi-object tracking algorithms employ \textit{detection-based tracking} \cite{Bergmann_2019,10.1007/978-3-319-11752-2_21, 10.1007/978-3-642-33709-3_25, article, 10.1007/978-3-642-15549-9_44, mul, LUO2021103448}---i.e., they first detect objects, and then associates detections across frames to form trajectories. Object detection is typically tackled by detectors such as Faster-RCNN \cite{he2016deep} or scale-dependent pooling (SDP) \cite{7780603}. The data association problem can be solving using either online or offline approaches. In offline approaches, all the frames are required and analyzed to associate detections \cite{9413215, 8820309, offmot, tang2016multiperson}, whereas
in online approaches, only adjacent frames are used \cite{hu2019joint, chen2018, wu2021track, shuai2021siammot}. For instance, \cite{2016} combines the Kalman Filter and Hungarian algorithm to effiicently associate objects.

\textbf{Uncertainty quantification.}
There has been recent interest in quantifying uncertainty for safety-critical applications of deep learning.
One approach is calibrated prediction, which calibrates prediction probabilities for classification \citep{guo2017calibration,park2021pac}, regression \citep{kuleshov2018accurate}, structured prediction \citep{kuleshov2015calibrated}, and reinforcement learning \citep{malik2019calibrated}, and more recently for object detection \citep{harakeh2021estimating}. However, these approaches do not provide guarantees.
PAC prediction sets provide such guarantees for individual models by predicting sets of labels (rather than individual labels) that contain the true label with high probability \citep{Park2020PAC}. We compose individual PAC prediction sets to obtain guarantees for object detection and tracking.


\section{Background on PAC Prediction Set}

We use the PAC prediction sets \citep{Park2020PAC} as a building block for our approach. Let $\Xs$ be the example space and $\Ys$ be the label space, concretely defined for detection and tracking. We let $D$ denote a distribution over $\Xs \times \Ys$. 

The goal of PAC prediction sets is to find a prediction set $C: \Xs \rightarrow 2^\Ys$ that contains the true label with high probability, while minimizing the expected prediction set size. In particular, the prediction set error $L_D(C)$ be $L_D(C) \coloneqq \Prob_{(x, y) \sim D}[ y \notin C(x) ]$. We want to find a prediction set $C$ that is approximately correct, \ie, given $\epsilon \in (0, 1)$, we want $L_D(c)\le\epsilon$. To construct $C$, we assume given a held-out calibration set of i.i.d. examples from $D$, \ie, $Z \sim D^n$. Due to the randomness in the calibration set, the prediction set may not always be approximately correct. Thus, we want to find a prediction set that is probably approximately correct (PAC), \ie, given $\epsilon,\delta\in(0,1)$, we want
\eqas{
	\Prob_{Z \sim D^n}\[ L_D(C) \le \epsilon \] \ge 1 - \delta. 
}
In this case, we say $C$ is $(\epsilon, \delta)$-correct.

We can trivially construct PAC $C$ by letting $C(x)=\Ys$ for all $x\in\Xs$. Thus, we additionally want to minimize the size of $C(x)$. To do so, \cite{Park2020PAC} proposes the following one-dimensional parameterization of prediction sets:
\eqas{
C_\tau(x) = \{ y \in \Ys \mid f(x, y) \ge \tau \},
}
where $\tau \in \realnum_{\ge 0}$ and $f:\Xs\times\Ys\to\mathbb{R}_{\ge0}$ is any given scoring function (e.g., the label probabilities output by a deep neural network). Then, they provide an algorithm that computes a parameter value $\tau$ that satisfies the PAC guarantee, \ie,
\eqa{
\hat\tau = \operatorname*{\arg\max}_{\tau \in \realnum_{\ge 0}}~ \tau ~~ \text{subj. to} \sum_{(x, y) \in Z} \mathbbm{1}\( y \notin C_\tau(x) \) \le k^*,
\label{eqn:algorithm}
}
where
\eqas{
k^* =\operatorname*{\arg\max}_{k\in\mathbb{N}\cup\{0\}}~k\qquad\text{subj. to}\qquad F(k;n,\epsilon) \le \delta,
}
$n=|Z|$, and $F(k; n, \epsilon)$ is the cumulative distribution function of $\text{Binomial}(n, \epsilon)$. Here, (\ref{eqn:algorithm}) returns the trivial solution $\hat\tau=0$ if the the optimization problem is infeasible. Maximizing $\tau$ corresponds to minimizing the prediction set size. Note that the achieved prediction set size depends on the quality of $f$; if $f$ outputs random scores, then (\ref{eqn:algorithm}) will produce the trivial solution $\hat\tau=0$. We have the following guarantee.
\begin{theorem}[\citep{Park2020PAC}] \label{thm:pred_set}
$C_{\hat\tau}$ is $(\epsilon, \delta)$-correct for $\hat\tau$ as in (\ref{eqn:algorithm}).
\end{theorem}


\section{PAC Object Detection}

\subsection{Problem}

Our goal is to design a prediction set for object detections that contains the true object location and class with high probability. We use a two-stage detector, Faster-RCNN, which consists of three components: proposal, location, and class presence. Let $x \in \Xs$ be an image, $b \in \Bs$ be a bounding box, $c \in \Ks$ be an object class, $e \in \{0, 1\}$ be a presence flag, and $y \coloneqq (b, c, e) \in \Ys \coloneqq \Bs \times \Ks \times \{0, 1\}$ be a pair of a bounding box and an object class along with a presence flag.
We let $D_\dtr$ denote the distribution over $\Xs \times \Ys$.
The two-stage detector generally assumes each proposal contains one instance per class, so we also make this assumption when designing prediction sets.
When matching a predicted bounding box to a ground truth one, we assume two bounding boxes are the same if the intersection over union (IoU) of the two boxes is larger than 0.25, following the usual convention in designing object detectors.

\subsection{Component-wise Prediction Sets} \label{sec:det_componentwisepredictionset}

Here, we describe how to construct prediction sets for each detection component (proposal, presence, and location) for given prediction set thresholds. In the next section, we describe how to find thresholds that provide PAC guarantees for individual components, and how to combine them to obtain an overall prediction set with PAC guarantees.

\textbf{Proposal prediction set.}
Given an image $x$, the proposal component $\fh_{\prp}: \Xs \rightarrow (\Bs \times \realnum_{\geq 0})^*$ outputs a set of pairs $(r,s)$ consisting of a bounding box (or \emph{proposal}) $r$ and an objectness score $s$, where $r$ is a region likely to contain an object and the $s$ is the likelihood that $r$ contains an object. Given a threshold $\tau_\prp$, we construct a prediction set that outputs bounding boxes that contain objects based on $\fh_{\prp}$:
\eqas{
	C_{\hat\tau_\prp}(x) \coloneqq \left\{ r \in \Bs \vmid (r, s) \in \fh_\prp(x) \wedge s \geq \hat\tau_\prp \right\}.
}
In other words, this prediction set contains bounding boxes from the proposal component with scores above $\tau_\prp$.

\textbf{Presence prediction set.}
Given an image $x$, a proposal $b$, and a target class $c$, the presence component $\fh_\prs: \Xs \times \Bs \times \Cs \rightarrow \realnum_{\ge 0}$ outputs the likelihood of the target class in the proposal. Given a threshold $\tau_\prs$, we construct a prediction set indicating whether the target class exists based on $\fh_\prs$:
\begin{multline*}
C_{\hat\tau_\prs}(x, r, c) \coloneqq \Big\{ e \in \{0, 1\} \;\Big|\; e \fh_\prs(x, r, c) + \\ (1-e)(1 - \fh_\prs(x, r, c)) \ge \hat\tau_\prs) \Big\},
\end{multline*}
where $e$ is a presence flag. The presence component $\fh_{\prs}$ outputs the probability of presence in the region $r$ if its class is $c$; the corresponding prediction set $\Ch_\prs$ represents whether $c$ exists in the region $r$. In the detector implementation, we may not have $\fh_\prs(x, r, c)$ for some $(x, r, c)$ due to the non-maximum suppression (more precisely, due to the smaller score $\fh_\prs(x, r, c)$ compared to the score of an adjacent bounding box $r'$). In this case, we treat it as if $\fh_\prs(x, r, c)$ outputs $0$ (i.e., it is always incorrect).

\textbf{Location prediction set.}
Given an image $x$, a proposal $r$, and a target class $c$, the location component $\fh_\loc: \Xs \times \Bs \times \Cs \rightarrow \Ps_\Bs$ outputs a distribution $\Ps_\Bs$ over bounding boxes $\Bs$. Given a threshold $\tau_\loc$, the location prediction set outputs all bounding boxes with target class object score above $\tau_\loc$:
\eqas{
	C_{\hat\tau_\loc}(x, r, c) \coloneqq \left\{ b \in \Bs \vmid \fh_\loc(b \mid x, r, c) \ge \hat\tau_\loc \right\}.
}
Intuitively, the location component $\fh_\loc$ outputs the probability of an object located in the region $r$ if its class is $c$, and the corresponding prediction set $C_{\hat\tau_\loc}$ contains all bounding boxes that include objects of class $c$ with high probability.

\textbf{Detection prediction set.}
The prediction set for the entire detector consists of the prediction sets for each detector component. We denote the detection prediction set as:
\eqas{
	\Ch_{\dtr}(x) \! \coloneqq \!\! \bigcup_{{r \in C_{\hat\tau_\prp}(x), c \in \Ks}, {e \in C_{\hat\tau_\prs}(x, r, c), b \in C_{\hat\tau_\loc} (x, r, c)}} \!\!\!\! \left\{ \(c, e, b \) \right\}.
}
For each proposal from the proposal prediction set $r$ and each target class $c$, the detector prediction set combines the location and presence prediction sets and outputs a collection of a presence flag and a location pair $(e, b)$.

\subsection{Compositional PAC Guarantee for Detection} \label{sec:pac_ps_det_composition}

Next, we describe how to compute parameters $\hat\tau_\prp$, $\hat\tau_\prs$, and $\hat\tau_\loc$ of detector prediction set such that $\Ch_{\det}$ is $(\epsilon, \delta)$-correct, while minimizing the prediction set size. In particular, given an i.i.d. calibration set $Z$, we provide a compositional PAC guarantee for the entire detection pipeline based on the component-wise PAC guarantees.

\textbf{Proposal prediction set.} First, we construct the PAC proposal prediction set by choosing $\hat\tau_\prp$ such that $C_{\hat\tau_\prp}$ is $(\epsilon_\prp, \delta_\prp)$-correct by using the algorithm encoded by (\ref{eqn:algorithm}). The prediction set error is high if the true bounding box is not in the proposal prediction set. In particular, the prediction set error is defined as follows:
\begin{align*}
	L_{D_\dtr}(C_{\hat\tau_\prp}) \coloneqq \Exp_{(x, y)}~ \ell_{\prp}^{01}(C_{\hat\tau_\prp}, x, y),
\end{align*}
where $\ell_{\prp}^{01}(\cdot) \coloneqq \mathbbm{1}( b \notin C_{\hat\tau_{\prp}}(x) )$.
In practice, one ground truth bounding box $b$ can be matched to multiple proposals $r$s---\ie, $\text{IoU}(b, r) \ge 0.25$. In this , we choose the $r$ with the smallest score $s$ when constructing the prediction set.

\textbf{Presence and location prediction sets.}
The proposal prediction sets are the foundation for the presence and location prediction sets. In particular, given a proposal prediction set $C_{\hat\tau_\prp}$, we define the presence prediction set error to be:
\eqas{
	L_{D_\dtr}(C_{\hat\tau_\prs}; C_{\hat\tau_\prp}) \coloneqq \Exp_{(x, y)} \ell_\prs^{01}(C_{\hat\tau_\prs}, C_{\hat\tau_\prp}, x, y),
}
where
\eqas{
	\ell_\prs^{01} (\cdot) \coloneqq \mathbbm{1}\( e \notin \bigcup_{\bh \in C_{\hat\tau_\prp}(x), \bh = b } C_{\hat\tau_\prs}(x, \bh, c)  \).
}
Importantly, we use the ground truth object proposer $C^*_\prp$ (\ie, output the ground truth bounding boxes) to estimate the presence prediction set $C_{\hat\tau_\prs}$.
The following theorem shows that this strategy ensures correctness for $C_{\hat\tau_\prs}$ even when it is used in conjunction with the estimated proposer $C_{\hat\tau_\prp}$.
\begin{theorem} \label{thm:comp_prp_prs}
If $C_{\hat\tau_\prp}$ is $(\epsilon_\prp, \delta_\prp)$-correct, and estimating $C_{\hat\tau_\prs}$ with $C^*_\prp$ is $(\epsilon_\prs, \delta_\prs)$-correct, then estimating the $C_{\hat\tau_\prs}$ based on $C_{\hat\tau_\prp}$ is $(\epsilon_\prp+\epsilon_\prs, \delta_\prp + \delta_\prs)$-correct.
\end{theorem}
See Section \ref{apdx:thm:comp_prp_prs_proof} for a proof.
Next, given a proposal prediction set $C_{\hat\tau_\prp}$, we define the error of the location prediction set:
\eqas{
	L_{D_\dtr}(C_{\hat\tau_\loc}; C_{\hat\tau_\prp}) \coloneqq \Exp_{(x, y)} \ell_\loc^{01}(C_{\hat\tau_\loc}, C_{\hat\tau_\prp}, x, y),
}
where
\eqas{
	\ell_\loc^{01} (\cdot) \coloneqq \mathbbm{1}\( b \notin \bigcup_{\bh \in C_{\hat\tau_\prp}(x), \bh = b } C_{\hat\tau_\loc}(x, \bh, c)  \).
}
The following theorem guarantees that the location prediction set is correct by composing the PAC proposal prediction set and PAC location prediction set with the true proposal prediction set; see Section \ref{apdx:thm:comp_prp_loc_proof} for a proof.
\begin{theorem} \label{thm:comp_prp_loc}
If $C_{\hat\tau_\prp}$ is $(\epsilon_\prp, \delta_\prp)$-correct and estimating $C_{\hat\tau_\loc}$ with $C^*_\prp$ is $(\epsilon_\loc, \delta_\loc)$-correct, then estimating $C_{\hat\tau_\loc}$ based on $C_{\hat\tau_\prp}$ is $(\epsilon_\prp+\epsilon_\loc, \delta_\prp + \delta_\loc)$-correct.
\end{theorem}

\textbf{Composed detection prediction set.}
The final detection prediction set combines three prediction sets: the proposal, presence, and location prediction sets. In particular, the detection prediction set error is defined as follows:
\eqas{
L_{{D_\dtr}}(\Ch_\dtr; C_{\hat\tau_\prs}, C_{\hat\tau_\loc}) \coloneqq \Expop_{(x, y)}\[ \ell_{\dtr}^{01}(\Ch_{\dtr}, C_{\hat\tau_\prs}, C_{\hat\tau_\loc}, x, y) \], 
}
where
\eqas{
\ell_{\dtr}^{01}(\cdot) \coloneqq \mathbbm{1} \( y \notin \Ch_{\dtr}(x; C_{\hat\tau_\prs}, C_{\hat\tau_\loc}) \).
}
To obtain an $(\epsilon_\dtr, \delta_\dtr)$-correct detector prediction set, we need an $(\epsilon_\prs', \delta_\prs')$-correct presence prediction set and an $(\epsilon_\loc', \delta_\loc')$-correct location prediction set;
here 
$\epsilon_\prs' = \epsilon_\prs + \epsilon_\prp$ and
$\delta_\prs' = \delta_\prs + \delta_\prp$ by Theorem \ref{thm:comp_prp_prs}, and $\epsilon_\loc' = \epsilon_\loc + \epsilon_\prp$ and
$\delta_\loc' = \delta_\loc + \delta_\prp$ by Theorem \ref{thm:comp_prp_loc}.
Then, the following theorem says the composed detection prediction set satisfies the desired $(\epsilon_\dtr, \delta_\dtr)$ guarantee, where $\epsilon_\dtr = \epsilon_\prs' + \epsilon_\loc'$ and $\delta_\dtr = \delta_\prs' + \delta_\loc'$ (see Section \ref{apdx:thm:comp_det_proof} for a proof):
\begin{theorem} \label{thm:comp_det}
If $C_{\hat\tau_\prs}$ is $(\epsilon_\prs', \delta_\prs')$-correct and
$C_{\hat\tau_\loc}$ is $(\epsilon_\loc', \delta_\loc')$-correct, $\Ch_\dtr$ is $(\epsilon_\prs'+\epsilon_\loc', \delta_\prs'+\delta_\loc')$-correct.
\end{theorem}


\section{PAC Multi-object Tracking}

\subsection{Problem}

PAC multi-object tracking aims to design a prediction set that contains an object trajectory with high probability. Given detection prediction sets for each frame, we want to model the object transition between two frames with the PAC guarantee. To this end, we define an \textit{edge prediction set} that contains the true object transition between frames with high probability. As before, let $x \in \Xs$ be an image, $b \in \Bs$ be a bounding box, $c \in \Ks$ be an object class, $e \in \{0, 1\}$ be a presence flag, and $y \coloneqq (b, c, e) \in \Ys \coloneqq \Bs \times \Ks \times \{0, 1\}$. In additional, let $x_{t}$ be an image at time step $t$, $y_t$ be a detection at time step $t$, and $(y_t, y_{t+1})$ be an edge between two detections. We construct edge prediction sets by combining the detection prediction set at each time step.

\subsection{Component-wise Prediction Sets}
We use a component-wise approach, \ie, independently designing a detection prediction set and edge prediction set for the composed prediction set for tracking.

\textbf{Detection prediction set.} We adopt the detection prediction set $\Ch_\dtr$ as defined in Section \ref{sec:det_componentwisepredictionset}.

\textbf{Edge prediction set.} Given two adjacent images $x_t$ and $x_{t+1}$ and their corresponding detection prediction sets $\Ch_\dtr(x_t)$ and $\Ch_\dtr(x_{t+1})$, an edge score function $\fh_\edge: \Xs^2 \times \Ys^2 \rightarrow \realnum$ computes the similarity between these detection prediction sets, \ie, $\yh_t \in \Ch_\dtr(x_{t})$ and $\yh_{t+1} \in \Ch_\dtr(x_{t+1})$ have high score if they are similar to each other. Given an edge score function and a threshold $\hat\tau_\edge$, we define the edge prediction set to include edges with score above $\hat\tau_\edge$:
\begin{multline*}
C_{\hat\tau_{\edge}}(x_t, x_{t+1}) \coloneqq 
\Big\{ (y_t, y_{t+1}) \in \Ch_{\det}(x_{t}) \times \Ch_{\det}(x_{t+1}) \\ \;\Big|\; \fh_\edge(x_t, x_{t+1}, y_t, y_{t+1}) \ge \hat\tau_\edge \Big\}.
\end{multline*}

\textbf{Edge score function.} We can use an arbitrary edge score function; as before, better score functions result in smaller prediction sets. For simplicity, we use the intersection over union (IoU) between two bounding boxes. In particular, we compute the IoU of two bounding boxes if their object class is the same and they are present, \ie, 
\eqas{
\fh_\edge(\cdot) \coloneqq 
\begin{cases}
\text{IoU}(b_t, b_{t+1}) &\text{~if~} c_t = c_{t+1} \wedge e_t=e_{t+1}=1 \\
0 &\text{~otherwise}.
\end{cases}
}

\subsection{Compositional PAC Guarantee for Tracking}
We describe the compositional approach for PAC Tracking based on the component-wise PAC guarantee. 

\textbf{Detection prediction set.} We assume given an $(\epsilon_\dtr, \delta_\dtr)$-correct detection prediction set, \eg, one constructed using the approach in Section \ref{sec:pac_ps_det_composition}.

\textbf{Edge prediction set.}
The edge prediction set $C_{\hat\tau_\edge}$ is built upon a detection prediction set $\Ch_\dtr$. Given the detection prediction set, the edge prediction set error is defined as:
\begin{multline*}
L_{D_\edge}(C_{\hat\tau_\edge}; \Ch_{\det}) \coloneqq \\ \Expop_{x_t, x_{t+1}, y_t, y_{t+1}}[ \ell_{\edge}^{01}(\cdot)C_{\hat\tau_\edge}, x_t, x_{t+1}, y_t, y_{t+1}) ],
\end{multline*}
where $D_\edge$ is a distribution over $\Xs^2 \times \Ys^2$ and
\eqas{
\ell_{\edge}^{01} (\cdot) \coloneqq \mathbbm{1}\( (y_t, y_{t+1}) \notin C_{\hat\tau_\edge}(x_t, x_{t+1}) \).
}

Given the true detection prediction set $C^*_\dtr$ (\ie, the prediction set that returns true detections), $\Ch_\edge$ is constructed to be $(\epsilon_\edge, \delta_\edge)$-correct.
Intuitively, the edge prediction set is reduced to the prediction set for multi-class classification.

Note that $C_{\hat\tau_\edge}$ is constructed using the true detection prediction set $C^*_\dtr$, which does not account for the error of the estimated detection prediction set $\Ch_\dtr$; the following theorem says that using $C_{\hat\tau_\edge}$ in conjunction with the estimated detection prediction set $\Ch_\dtr$ still satisfies the PAC guarantee based on the edge prediction sets with the true detection prediction set $C^*_\dtr$; see Appendix \ref{apdx:thm:det_edge_proof} for a proof.

\begin{theorem} \label{thm:edge_composed}
If 
$\Ch_{\dtr}$ is $(\epsilon_\dtr, \delta_\dtr)$-correct and 
$C_{\hat\tau_\edge}$, which is based on $C^*_\dtr$, is $(\epsilon_\edge, \delta_\edge)$-correct, then
 estimating $C_{\hat\tau_\edge}$ using $\Ch_\dtr$ is $(\epsilon_\dtr+\epsilon_\edge, \delta_\dtr+\delta_\edge)$-correct.
\end{theorem}

\section{Experiments}

\subsection{Object Detection}

\begin{figure}
\centering
\includegraphics[width=0.8\linewidth]{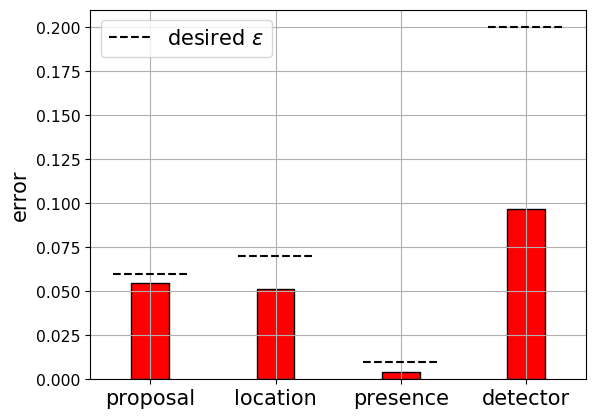}
\caption{Prediction set errors. Parameters are $\epsilon_\dtr=0.2$ and $\delta_\dtr = 10^{-5}$. The component-wise and composed prediction sets satisfy the desired prediction set error rate (\ie, the empirical errors (the red bars) are below the desired error rates (the dotted lines).}
\label{fig:ps_iid_det_error}
\end{figure}

\begin{figure*}[bt!]
\centering
\begin{subfigure}[b]{0.24\linewidth}
\centering
\includegraphics[width=0.7\linewidth]{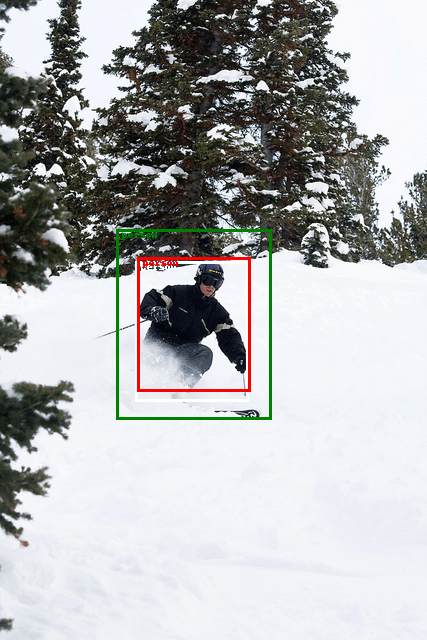} 
\end{subfigure}
\begin{subfigure}[b]{0.24\linewidth}
\centering
\includegraphics[width=\linewidth]{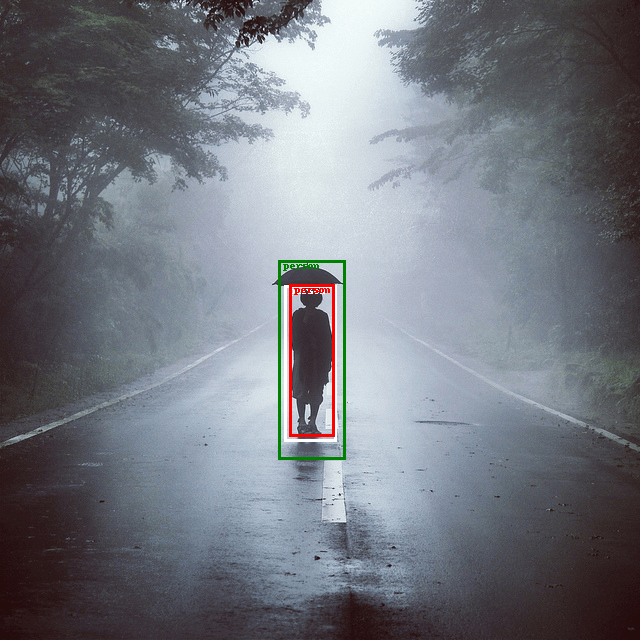} 
\end{subfigure}
\begin{subfigure}[b]{0.24\linewidth}
\centering
\includegraphics[width=\linewidth]{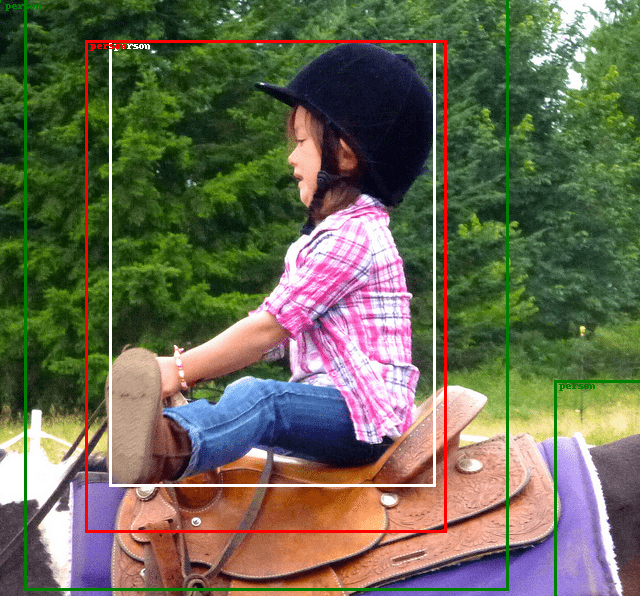} 
\end{subfigure}
\begin{subfigure}[b]{0.24\linewidth}
\centering
\includegraphics[width=\linewidth]{./figs/coco/ps_ex_eps_0.2/25.png} 
\end{subfigure}

\begin{subfigure}[b]{0.24\linewidth}
\centering
\includegraphics[width=\linewidth]{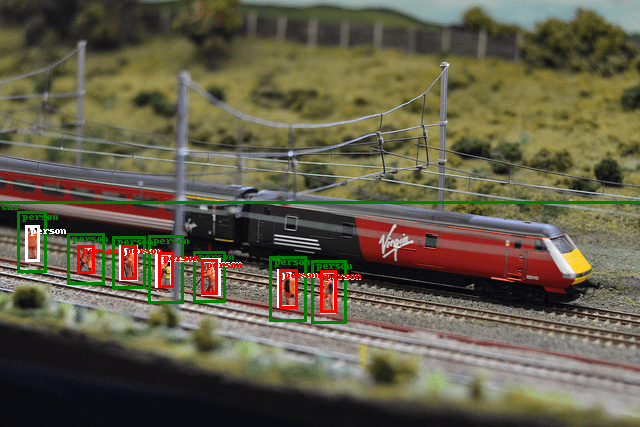} 
\end{subfigure}
\begin{subfigure}[b]{0.24\linewidth}
\centering
\includegraphics[width=\linewidth]{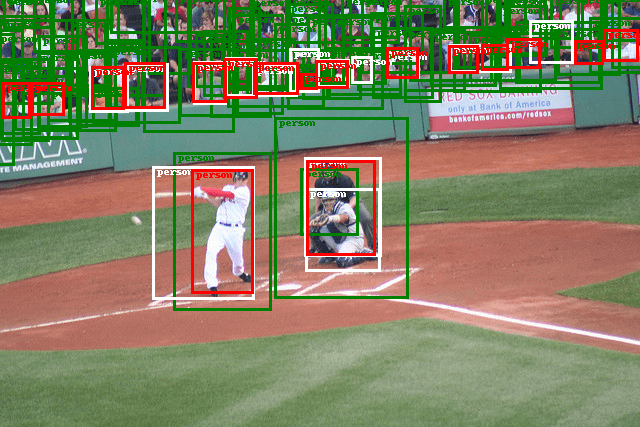} 
\end{subfigure}
\begin{subfigure}[b]{0.24\linewidth}
\centering
\includegraphics[width=\linewidth]{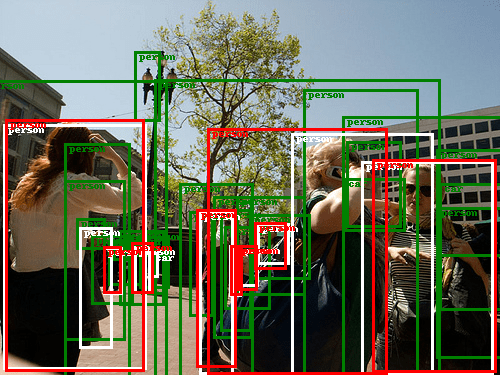} 
\end{subfigure}
\begin{subfigure}[b]{0.24\linewidth}
\centering
\includegraphics[width=\linewidth]{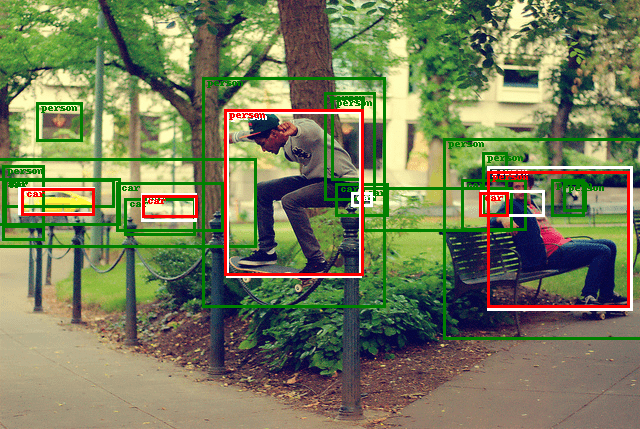} 
\end{subfigure}

\begin{subfigure}[b]{0.24\linewidth}
\centering
\includegraphics[width=\linewidth]{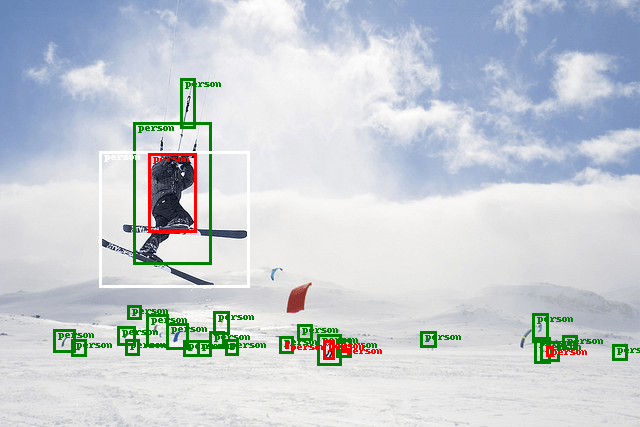} 
\end{subfigure}
\begin{subfigure}[b]{0.24\linewidth}
\centering
\includegraphics[width=\linewidth]{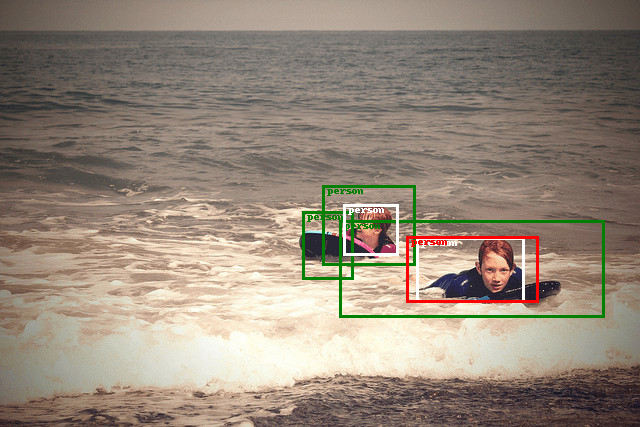} 
\end{subfigure}
\begin{subfigure}[b]{0.24\linewidth}
\centering
\includegraphics[width=\linewidth]{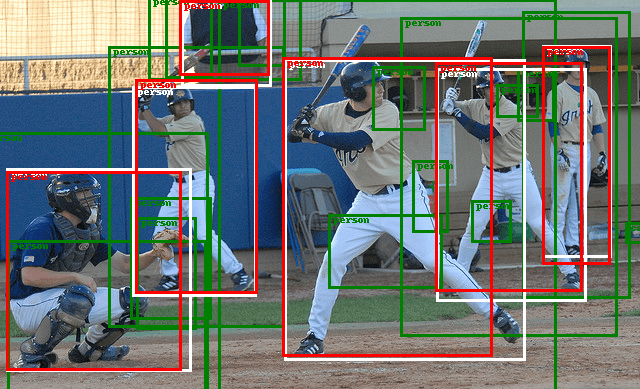} 
\end{subfigure}
\begin{subfigure}[b]{0.24\linewidth}
\centering
\includegraphics[width=\linewidth]{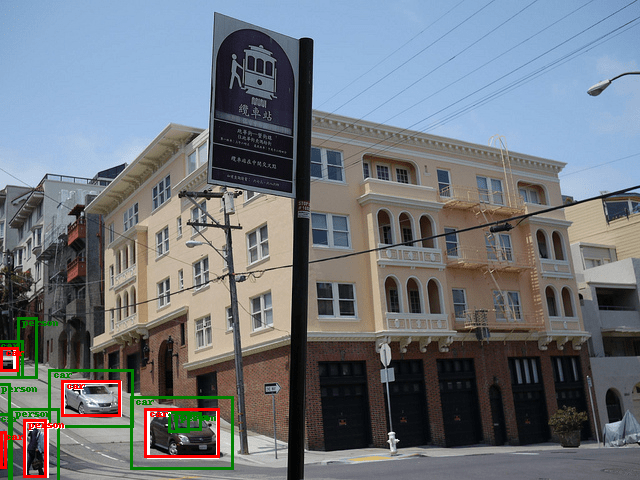} 
\end{subfigure}


\begin{subfigure}[b]{0.24\linewidth}
\centering
\includegraphics[width=\linewidth]{./figs/coco/ps_ex_eps_0.2/4.png} 
\end{subfigure}
\begin{subfigure}[b]{0.24\linewidth}
\centering
\includegraphics[width=\linewidth]{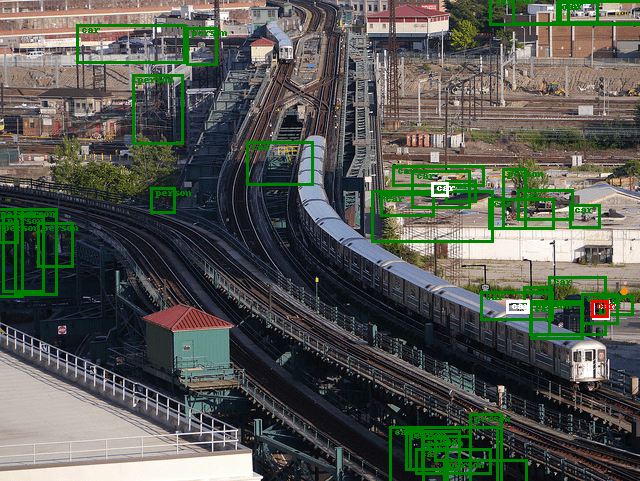} 
\end{subfigure}
\begin{subfigure}[b]{0.24\linewidth}
\centering
\includegraphics[width=\linewidth]{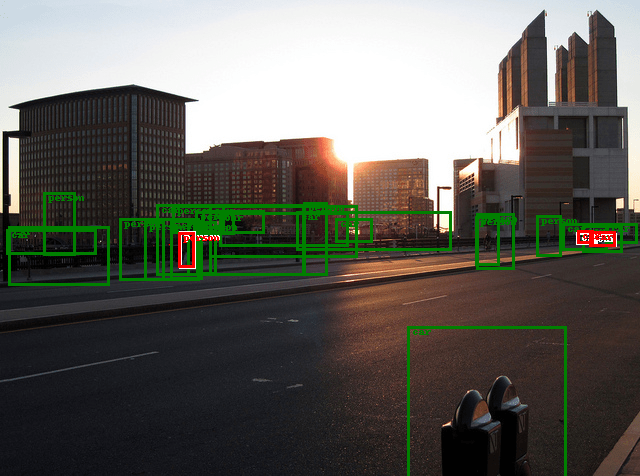} 
\end{subfigure}
\begin{subfigure}[b]{0.24\linewidth}
\centering
\includegraphics[width=\linewidth]{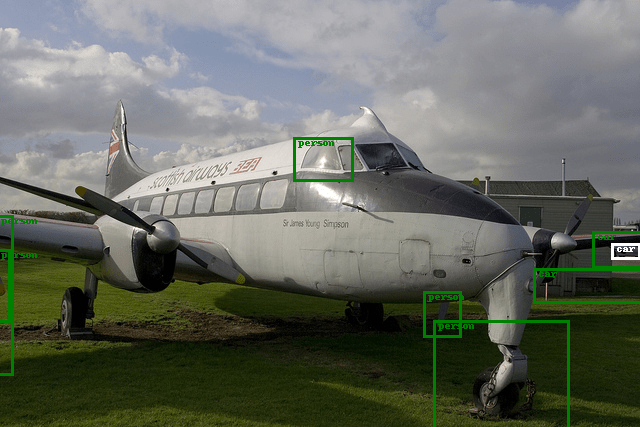} 
\end{subfigure}
\caption{Examples of prediction sets on the COCO dataset. Parameters are $\epsilon_\dtr=0.15$ and $\delta_\dtr = 10^{-5}$. White bounding boxes are ground truth, red bounding boxes are Faster RCNN predictions with class scores greater than 0.5, and green bounding boxes are prediction sets.}
\label{fig:ps_iid_det_coco}
\end{figure*}

\begin{figure*}[bt!]
\centering
\begin{subfigure}[b]{0.32\linewidth}
\centering
\includegraphics[width=\linewidth]{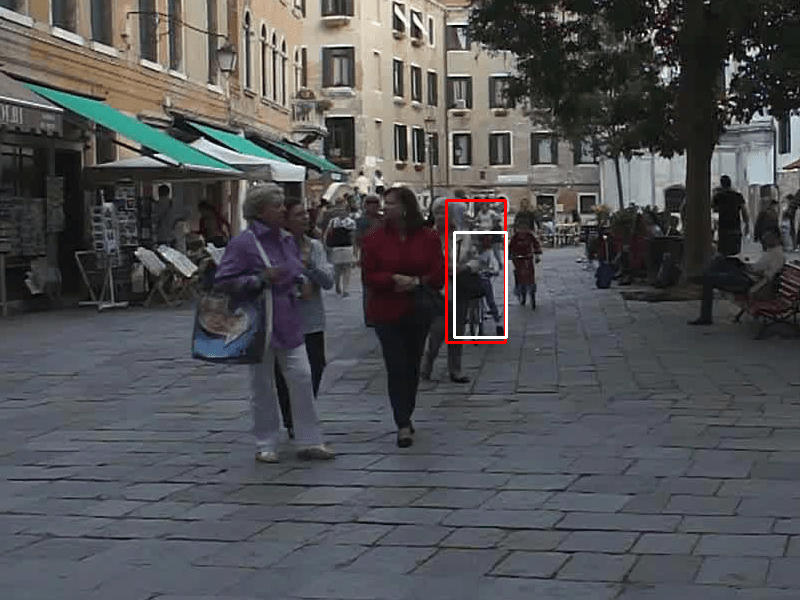} 
\end{subfigure}
\begin{subfigure}[b]{0.32\linewidth}
\centering
\includegraphics[width=\linewidth]{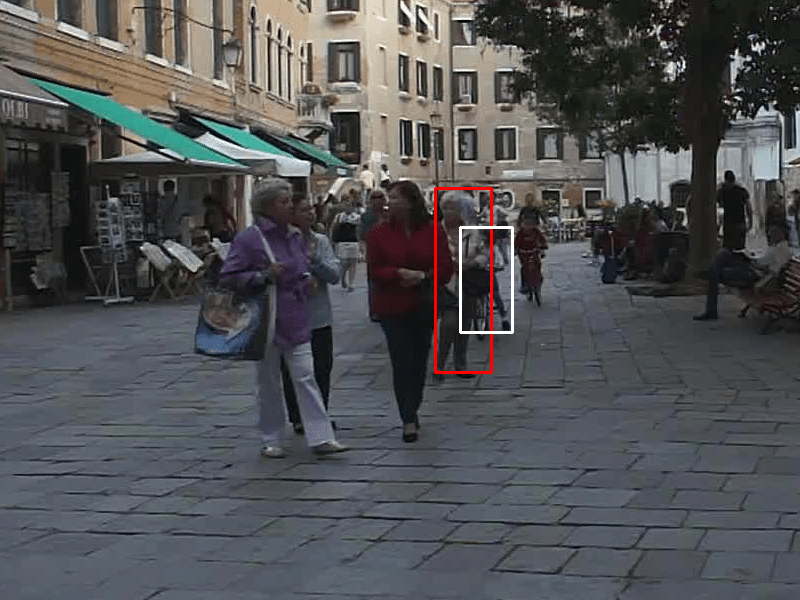} 
\end{subfigure}
\begin{subfigure}[b]{0.32\linewidth}
\centering
\includegraphics[width=\linewidth]{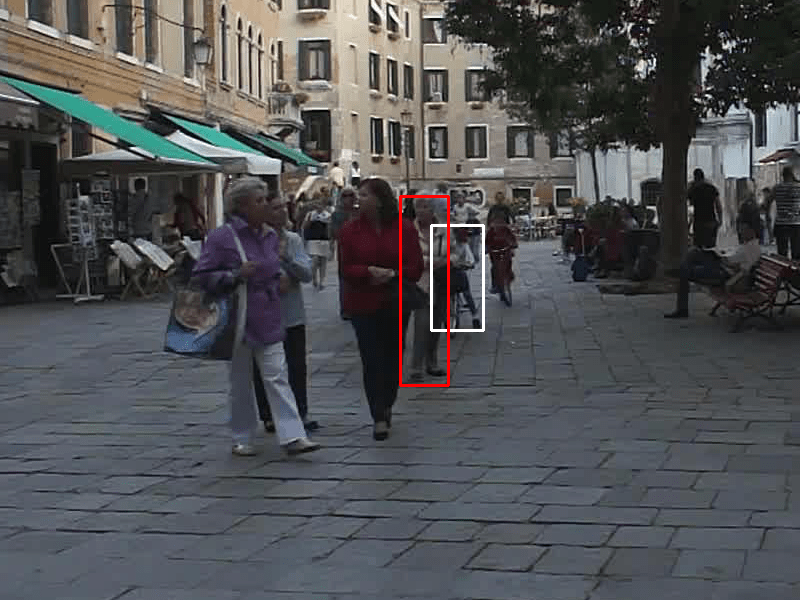} 
\end{subfigure}

\begin{subfigure}[b]{0.32\linewidth}
\centering
\includegraphics[width=\linewidth]{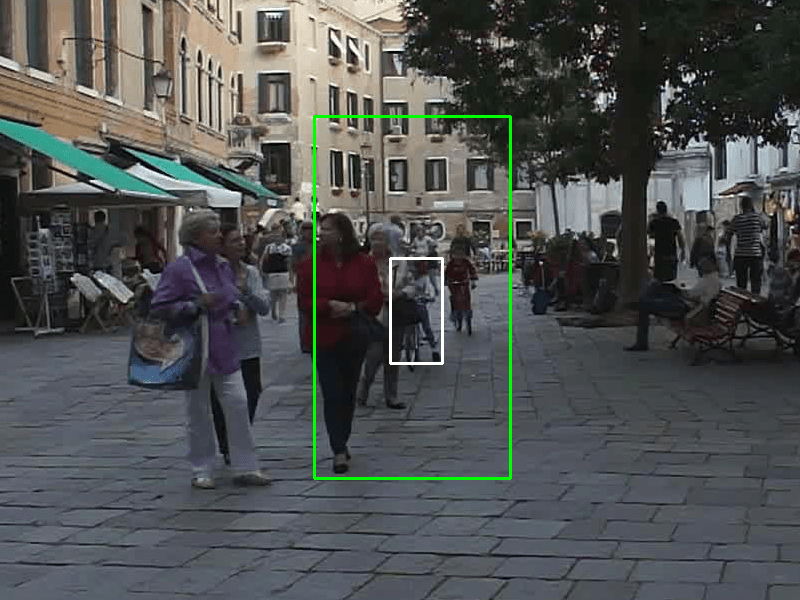} 
\end{subfigure}
\begin{subfigure}[b]{0.32\linewidth}
\centering
\includegraphics[width=\linewidth]{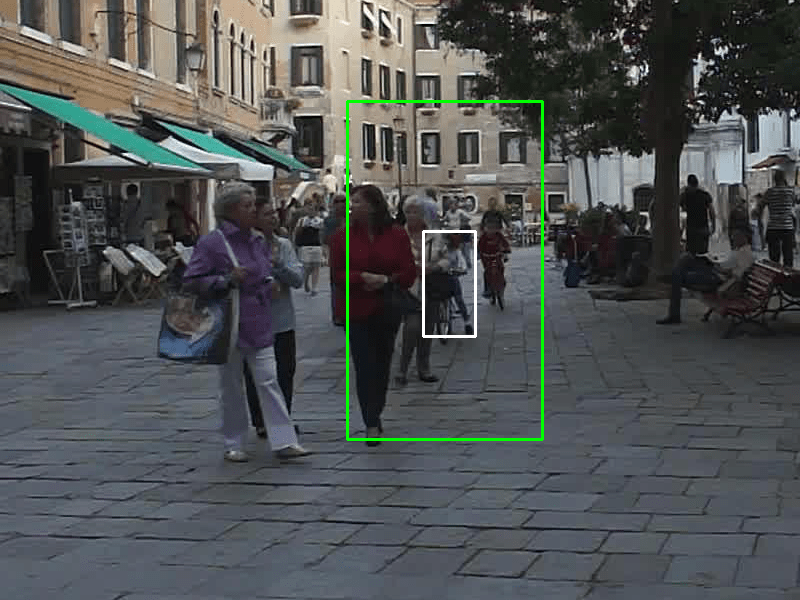} 
\end{subfigure}
\begin{subfigure}[b]{0.32\linewidth}
\centering
\includegraphics[width=\linewidth]{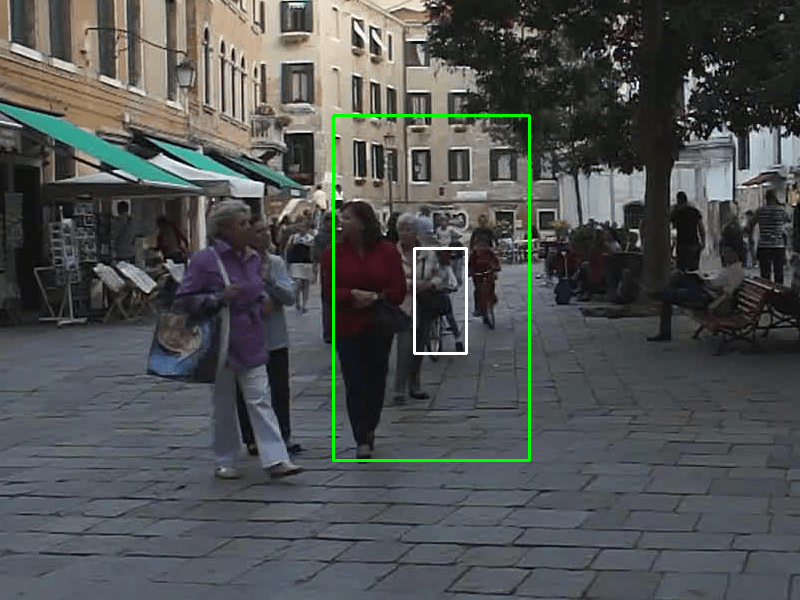} 
\end{subfigure}



\vspace{10pt}

\begin{subfigure}[b]{0.32\linewidth}
\centering
\includegraphics[width=\linewidth]{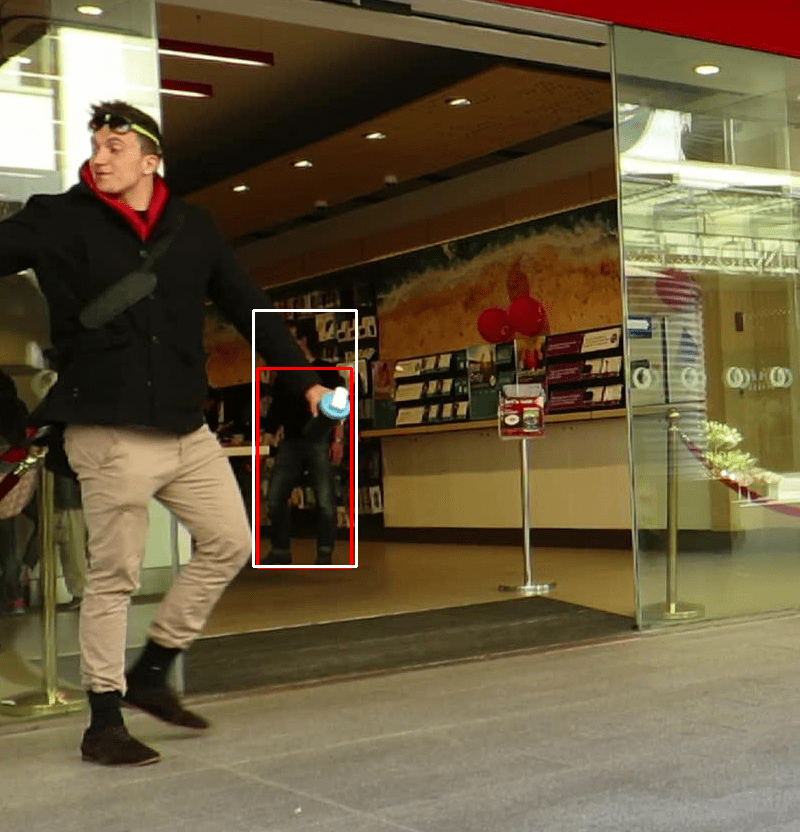} 
\end{subfigure}
\begin{subfigure}[b]{0.32\linewidth}
\centering
\includegraphics[width=\linewidth]{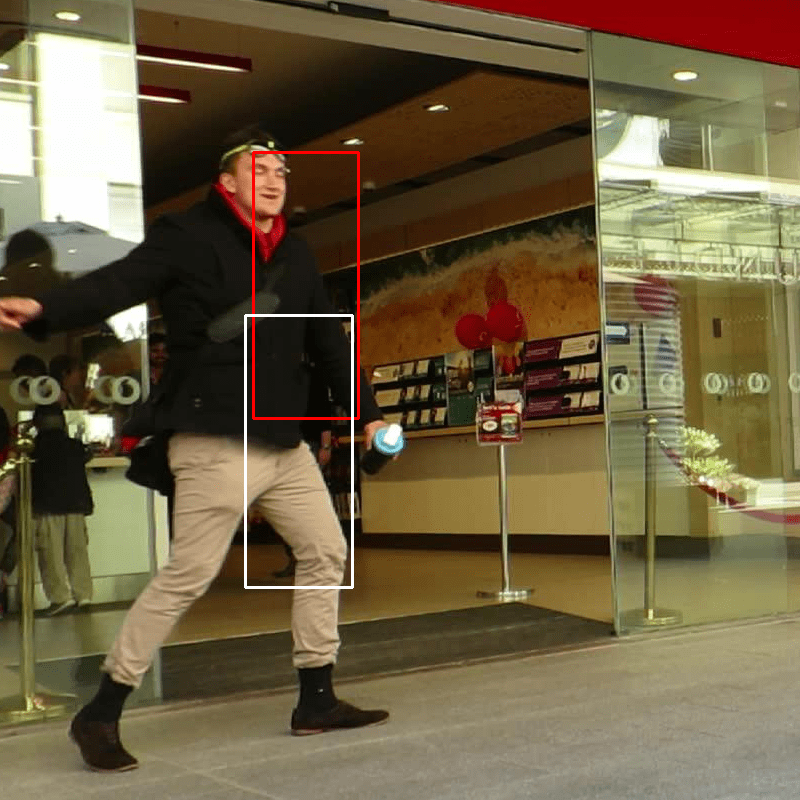} 
\end{subfigure}
\begin{subfigure}[b]{0.32\linewidth}
\centering
\includegraphics[width=\linewidth]{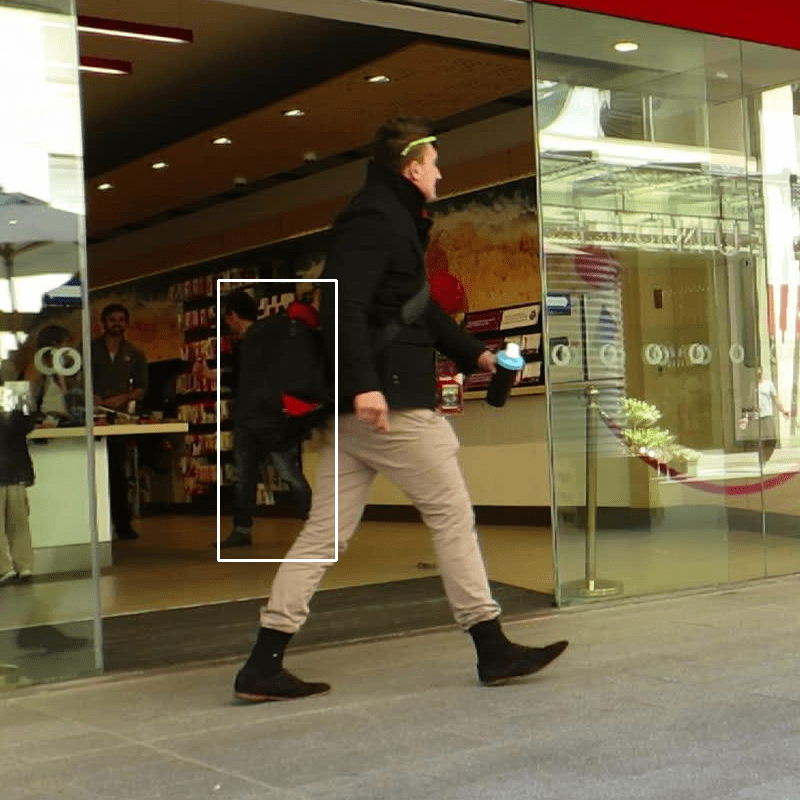} 
\end{subfigure}
\begin{subfigure}[b]{0.32\linewidth}
\centering
\includegraphics[width=\linewidth]{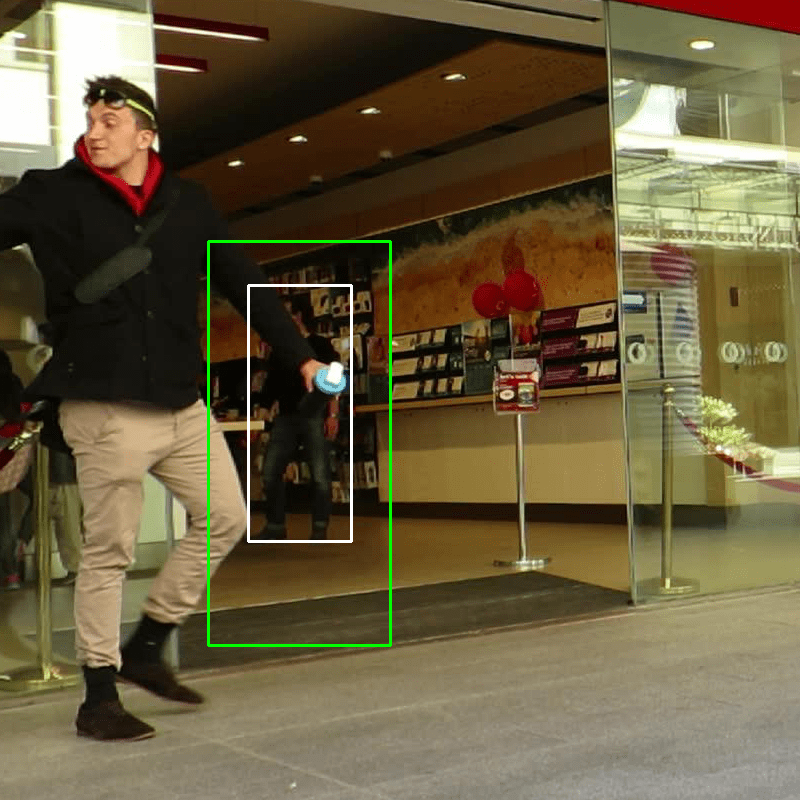} 
\end{subfigure}
\begin{subfigure}[b]{0.32\linewidth}
\centering
\includegraphics[width=\linewidth]{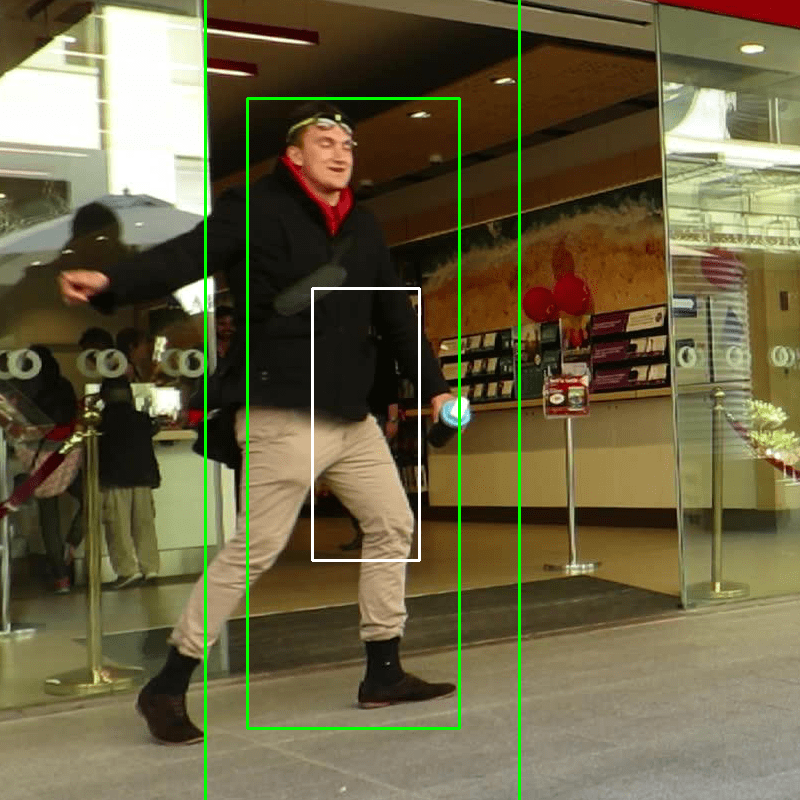} 
\end{subfigure}
\begin{subfigure}[b]{0.32\linewidth}
\centering
\includegraphics[width=\linewidth]{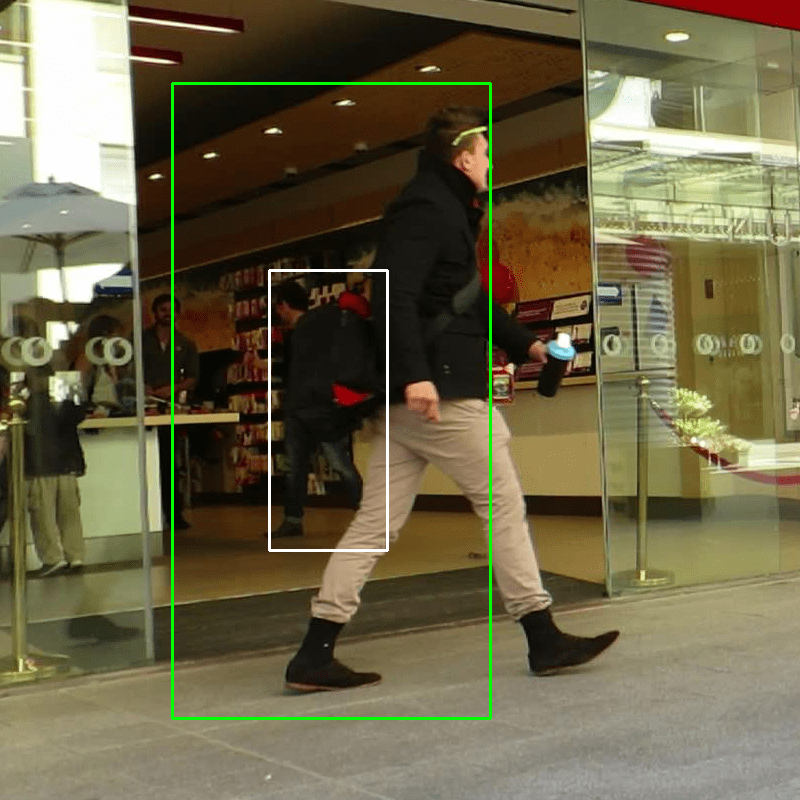} 
\end{subfigure}
\caption{Composition of detection prediction set and edge prediction sets. White boxes are ground truth, red boxes are bounding boxes of tracks in Tracktor's detection, and green boxes are our edge prediction sets $C_{\hat\tau_\edge}$. Rows 1 and 3 are detection and tracking results from Tracktor, and rows 2 and 4 are our results. In these cases, our detection prediction sets contain the ground truth object bounding boxes with high probability, and our edge prediction sets contain the track that tracks our interested object with high probability, but Tracktor does not.}
\label{fig:ps_comp}
\end{figure*}

We apply the proposed compositional PAC prediction set approach in Section \ref{sec:pac_ps_det_composition} over the faster RCNN object detector \citep{ren2015faster} and evaluate it using the COCO dataset \citep{lin2014microsoft} with the labels ``person'' and ``car''.
In particular, the COCO dataset consists of $117,266$ training images, $2,476$ calibration images, and $2,476$ test images, where we have split the original test images into calibration and test images. We construct the prediction sets in a compositional way with the detector parameters $\epsilon_\dtr = 0.2$ and $\delta_\dtr = 10^{-5}$. Based on Theorem \ref{thm:comp_prp_loc}, we set up the proposal, presence, and location prediction sets with the following parameters: $\epsilon_\prp = 0.03$, $\epsilon_\prs=0.01$, $\epsilon_\loc=0.06$, and $\delta_\prp = \delta_\prs = \delta_\loc = 1/3 \times 10^{-5}$. The numbers of calibration examples for proposal, presence, and location prediction sets are $6,565$, $6,033$, and $6,033$ respectively.
Note that the object proposer $\fh_\prp$ only outputs a finite number of proposals in practice, so it introduces irreducible prediction set error even with $\tau_\prp = 0$. Thus, $\epsilon_\prp$ cannot be arbitrarily small even with large $n$.

\textbf{Quantitative results.} Figure \ref{fig:ps_iid_det_error} shows the empirical prediction set error; the composed detection prediction set error is $0.097$, which is less than the desired error $\epsilon_\dtr = 0.2$. Also, each component of the composed detector \ie, proposal, location and presence, satisfy the desired error rates. 

\textbf{Qualitative results.} Figure \ref{fig:ps_iid_det_coco} visualizes the proposed prediction sets in green along with ground truth bounding boxes in white and the original detection results in red (by thresholding the class scores by $0.5$).
The first image shows that the prediction set includes ``person'' while having smaller bounding box, as desired. 
The fifth image shows that ``person'' prediction sets contains all ``person'' objects, whereas the original detection results miss one ``person''.
The images in the last row demonstrate the conservative output of prediction sets; in particular, the prediction set can control false negatives by the specified level of $\epsilon_\dtr$, but it introduces a few false positives; as seen in the images, the prediction set does not contain any actual objects. However, in safety-critical environments, controlling the false negative rate by allowing a few false positives is a worthwhile tradeoff to ensure safety (\eg, do not hit any ``person'' objects).

\subsection{Object Tracking}
We evaluate the edge prediction set method on the MOT17 dataset. In particular, we use ground truth sequences 02, 04, 05, 09, 10, and 11 in MOT17 and split each sequence into two halves, where the first half is used for a calibration set to derive the PAC threshold and the rest is used for the testing set. Then, we evaluate whether the edge prediction set satisfies the desired PAC guarantee. In our experiments, we use $\epsilon_{edge}=0.005$ and $\delta_{edge}=10^{-2}$, and evaluate our method using the false-negative rate (FNR) and average false-positive (AFP). The false-negative rate is defined to be the average rate at which true bounding boxes are not in the edge prediction set across time steps for all sequences, \ie,
\eqas{
	\text{FNR} = \frac{1}{|Z_\text{eval}|} \sum_{(x, x', y, y') \in Z_\text{eval}} \ell_{\edge}^{01}(\cdot),
}
where
\begin{multline*}
Z_{\text{eval}} = \big\{(x_t^{(s)}, x_{t+1}^{(s)}, y_t^{(s, i)}, y_{t+1}^{(s, i)}) \mid 
s \in S,~ 1 \le i \le O_t^{(s)}, \\
1 \le t \le T_s - 1
\big\},
\end{multline*}
$S$ is a set of a calibration part of each ground trouth sequences, 
$T_s$ is the number of time steps in a sequence $s$, and
$O_t^{(s)}$ is the set of objects in the time step $t$ of the sequence $s$.
The average false-positive is defined to be the average rate at which incorrect bounding boxes are included in the edge prediction set, \ie,
\begin{multline*}
\text{AFP} = \frac{1}{|Z_\text{eval}|} \sum_{(x, x', y, y') \in Z_\text{eval}} \big[	| C_{\hat\tau_\edge}(x, x')| \\
- \mathbbm{1}((y, y') \in C_{\hat\tau_\edge}(x, x')) \big].
\end{multline*}

\textbf{Quantitative results.}
We compare our method to a na\"{i}ve baseline where the edge prediction set includes bounding boxes with the top-1 to top-5 IoUs. As can be seen in Table \ref{tab:fn_fp_17}, our approach has the smallest false positive rates among approaches that satisfy the desired false negative rate $\epsilon_{edge}$;
In the second row of Table \ref{tab:fn_fp_17}, the top-1 set fails to satisfy the desired rate $\epsilon_{edge}=0.005$, whereas our PAC edge prediction set always satisfies this rate. Alternatively, the top-2 to top-5 prediction sets always satisfy the $\epsilon_{edge}=0.005$ rate, but they have worse false positive rates than our approach.

\begin{table}[t]
\centering
\begin{tabular}{c|cc}
\toprule
Prediction sets & FNR & AFP \\ \midrule
Edge     & 0.004              & \textbf{0.445}              \\ 
Top-1   & 0.012              & 0.012              \\ 
Top-2   & 0.004              & 0.502              \\ 
Top-3   & 0.004              & 0.668              \\ 
Top-4   & 0.004              & 0.751              \\ 
Top-5   & 0.004              & 0.801              \\ \bottomrule
\end{tabular}
\caption{Component-wise prediction set results in terms of false negative rates (FNR) and average false positive (AFP) for object tracking. Parameters are $\epsilon_{edge} = 0.005$ and $\delta_{edge} = 0.01$. Our edge prediction set always satisfied the desired $\epsilon_{edge} = 0.005$ FNR guarantee. In contrast, the top-1 set violates this constraint, whereas the top-2 to top-5 sets achieve worse AFP than our approach.}
\label{tab:fn_fp_17}
\end{table}

Table \ref{tab:edge_det_pred} shows that the composed edge prediction set empirically satisfies the desired false negative rate. In particular, the edge prediction set $C_{\hat\tau_\edge}$ is constructed using the true detection prediction set $C^*_\dtr$ to satisfy the desired $\epsilon_\edge$ false negative rate; using it in conjunction with the estimated detection prediction set $C_{\hat\tau_\dtr}$ continues satisfies the $\epsilon_\edge + \epsilon_\dtr$ false negative rates, as guaranteed by Theorem \ref{thm:edge_composed}.

\begin{table}[t]
\centering
\begin{tabular}{ccc|cc}
\toprule
$\epsilon_{det}$ & $\epsilon_{edge}$ & \makecell{desired FNR \\ $(\epsilon_{edge}+\epsilon_{det})$} & FNR & AFP \\ \midrule
0.2 & 0.01 & 0.210   & 0.136              & 0.324              \\ 
0.2 & 0.005 & 0.205     & 0.051              & 0.542              \\ 
0.2 & 0.001 & 0.201   & 0.023              & 0.738              \\  
\bottomrule
\end{tabular}
\caption{Composed prediction set results in terms of false negative rates (FNR) and average false positive (AFP) with various $\epsilon_{edge}$. In all three cases, the FNR is smaller than the desired $\epsilon_{edge}+\epsilon_{det}$.}
\label{tab:edge_det_pred}
\end{table}

\textbf{Qualitative results.}
Next, in Figure \ref{fig:ps_comp}, we show representative results for both the original Tracktor model as well as our method on the MOT17 dataset; here, white boxes are ground truth, red boxes are bounding boxes of tracks in Tracktor's detections,  and green boxes are our edge prediction set $C_{\hat\tau_\edge}$. Rows 1 and 3 are detection and tracking results from Tracktor, and rows 2 and 4 are results from our method. Furthermore, rows 1 and 2 are from \textit{sequence-02}, and rows 3 and 4 are from \textit{sequence-09}. We visualize the tracking results for tracking a single object in each sequence.

For the first example (\ie, rows 1 and 2), Tracktor's track (\ie, the red boxes in row 1) initially identifies the object of interest in the first image; however, it switches to another object in the second and third images, so it outputs the incorrect track.
In contrast, our edge prediction set (\ie, the. green boxes in row 2) include the ground truth track of the object (\ie, the green box contains the white box in all three images). The second example (\ie, rows 3 and 4) is similar. Specifically, Tracktor's track (\ie, the red boxes in row 3) deviate from the original object and even disappears in the third image (since Tracktor thinks it is not tracking anything meaningful, so it sets the track to inactive). In contrast, our method (\ie, green boxes in row 4) outputs a track that contains the object in all three images.

In addition, there are cases where the Tracktor bounding boxes do not contain ground truth bounding boxes, \eg, in the first image of row 3, the detected bounding box (red) is much smaller than the ground truth bounding box (white). In contrast, as shown on row 4, our detection prediction sets (green) always include the entire ground truth.

These examples demonstrate that our method detects and tracks objects whereas Tracktor fails. 
We provide another example in Appendix \ref{apdx:emp_aux}.



\section{Conclusion and Future Direction}


We have proposed a novel algorithm for constructing probably approximately correct (PAC) prediction sets for multi-object detection and tracking. For detection problem, our approach constructs PAC prediction sets for each component of an object detector based on uncertainty scores produced by the detector, and then composes them to form PAC prediction sets for the overall detector. For the tracker, our approach constructs edge prediction sets based on IoUs between bounding boxes at adjacent time steps in the video. In other words, with high probability, the prediction sets contain the true object and the edge prediction sets includes the true associates. We evaluate our method on COCO and MOT-17 dataset. Our results show that our prediction sets satisfy the desired $(\epsilon,\delta)$-correct property, and that our edge prediction set outperforms na\"{i}ve baselines.

One limitation of our method is that our prediction sets may be conservative. Since our approach must work in the worst-case scenario where the original detector performs poorly, our detection and edge prediction sets may be large in size, \ie, it may include false detections or overly large bounding boxes. Alternatively, it may introduce spurious edges in the edge prediction set (\eg, if there are many occlusions in the video) leading to spurious object trajectories.
However, we emphasize that large prediction sets represent errors or poor uncertainty scores in the underlying detection and tracking models; thus, to improve our prediction sets, it suffices to improve the underlying models used for detection and tracking. For instance, we could leverage the dynamics of objects to filter out unlikely tracks and shrink the sets. 

By applying a PAC guarantee over the detector and tracker in autonomous navigation, we enhance the trustworthiness of the system, which could contribute to mitigating the risk of pedestrian fatalities and bring about a positive societal impact.

{
    \small
    \bibliographystyle{ieee_fullname}
    \bibliography{macros,main}
}

\clearpage
\appendix
\onecolumn
\setcounter{page}{1}

\section{Proof of Theorem \ref{thm:comp_prp_prs}}
\label{apdx:thm:comp_prp_prs_proof}

Let 
\eqas{
	E_\prp &\coloneqq \{ (x, y) \in \Xs \times \Ys \mid \ell_\prp^{01}(\Ch_\prp, x, y) = 1 \},  \\
	U_\prs & \coloneqq \Xs \times \Ys \setminus E_\prp.
}
Then, the following holds:
\eqas{
	L_D(\Ch_\prs; \Ch_\prp) 
	&= \Exp_{x, y}\[ \ell_\prs^{01}(\Ch_\prs, \Ch_\prp, x, y) \] \\
	&= \int \ell_\prs^{01}(\Ch_\prs, \Ch_\prp, x, y) p(x, y)~\mathrm{d}x \mathrm{d}y \\
	&= \int_{E_\prp} \ell_\prs^{01}(\Ch_\prs, \Ch_\prp, x, y) p(x, y)~\mathrm{d}x \mathrm{d}y + \int_{U_\prs} \ell_\prs^{01}(\Ch_\prs, \Ch_\prp, x, y) p(x, y)~\mathrm{d}x \mathrm{d}y \\
	&\le \int_{E_\prp} p(x, y)~\mathrm{d}x \mathrm{d}y + \int_{U_\prs} \ell_\prs^{01}(\Ch_\prs, \Ch_\prp, x, y) p(x, y)~\mathrm{d}x \mathrm{d}y \\
	&= \int_{E_\prp} \ell_{\prp}^{01}(\Ch_\prp, x, y) p(x, y)~\mathrm{d}x \mathrm{d}y + \int_{U_\prs} \ell_\prs^{01}(\Ch_\prs, \Ch_\prp, x, y) p(x, y)~\mathrm{d}x \mathrm{d}y \\
	&= \int \ell_{\prp}^{01}(\Ch_\prp, x, y) p(x, y)~\mathrm{d}x \mathrm{d}y + \int_{U_\prs} \ell_\prs^{01}(\Ch_\prs, \Ch_\prp, x, y) p(x, y)~\mathrm{d}x \mathrm{d}y \\
	&= L_D(C_\prp) + \int_{U_\prs} \ell_\prs^{01}(\Ch_\prs, \Ch_\prp, x, y) p(x, y)~\mathrm{d}x \mathrm{d}y \\
	&= L_D(C_\prp) + \int_{U_\prs} \ell_\prs^{01}(\Ch_\prs, C_\prp^*, x, y) p(x, y)~\mathrm{d}x \mathrm{d}y \\
	&\le L_D(C_\prp) + \int \ell_\prs^{01}(\Ch_\prs, C_\prp^*, x, y) p(x, y)~\mathrm{d}x \mathrm{d}y \\
	&= L_D(C_\prp) + L_D(\Ch_\prs; C_\prp^*) \\
	&\le \epsilon_\prp + \epsilon_\prs,
}
where the last inequality holds with probability at least $1 - (\delta_p + \delta_l)$ due to the union bound.

\section{Proof of Theorem \ref{thm:comp_prp_loc}} \label{apdx:thm:comp_prp_loc_proof}
The proof is similar with that of Theorem \ref{thm:comp_prp_prs}.

Let 
\eqas{
	E_\prp &\coloneqq \{ (x, y) \in \Xs \times \Ys \mid \ell_\prp^{01}(\Ch_\prp, x, y) = 1 \},  \\
	U_\loc & \coloneqq \Xs \times \Ys \setminus E_\prp.
}
\eqas{
	L_D(\Ch_\loc; \Ch_\prp) 
	&= \Exp_{x, y}\[ \ell_\loc^{01}(\Ch_\loc, \Ch_\prp, x, y) \] \\
	&= \int \ell_\loc^{01}(\Ch_\loc, \Ch_\prp, x, y) p(x, y)~\mathrm{d}x \mathrm{d}y \\
	&= \int_{E_\prp} \ell_\loc^{01}(\Ch_\loc, \Ch_\prp, x, y) p(x, y)~\mathrm{d}x \mathrm{d}y + \int_{U_\loc} \ell_\loc^{01}(\Ch_\loc, \Ch_\prp, x, y) p(x, y)~\mathrm{d}x \mathrm{d}y \\
	&\le \int_{E_\prp} p(x, y)~\mathrm{d}x \mathrm{d}y + \int_{U_\loc} \ell_\loc^{01}(\Ch_\loc, \Ch_\prp, x, y) p(x, y)~\mathrm{d}x \mathrm{d}y \\
	&= \int_{E_\prp} \ell_{\prp}^{01}(\Ch_\prp, x, y) p(x, y)~\mathrm{d}x \mathrm{d}y + \int_{U_\loc} \ell_\loc^{01}(\Ch_\loc, \Ch_\prp, x, y) p(x, y)~\mathrm{d}x \mathrm{d}y \\
	&= \int \ell_{\prp}^{01}(\Ch_\prp, x, y) p(x, y)~\mathrm{d}x \mathrm{d}y + \int_{U_\loc} \ell_\loc^{01}(\Ch_\loc, \Ch_\prp, x, y) p(x, y)~\mathrm{d}x \mathrm{d}y \\
	&= L_D(C_\prp) + \int_{U_\loc} \ell_\loc^{01}(\Ch_\loc, \Ch_\prp, x, y) p(x, y)~\mathrm{d}x \mathrm{d}y \\
	&= L_D(C_\prp) + \int_{U_\loc} \ell_\loc^{01}(\Ch_\loc, C_\prp^*, x, y) p(x, y)~\mathrm{d}x \mathrm{d}y \\
	&\le L_D(C_\prp) + \int \ell_\loc^{01}(\Ch_\loc, C_\prp^*, x, y) p(x, y)~\mathrm{d}x \mathrm{d}y \\
	&= L_D(C_\prp) + L_D(\Ch_\loc; C_\prp^*) \\
	&\le \epsilon_\prp + \epsilon_\loc,
}
where the last inequality holds with probability at least $1 - (\delta_\prp + \delta_\loc)$ due to the union bound.

\section{Proof of Theorem \ref{thm:comp_det}} \label{apdx:thm:comp_det_proof}

Let
\eqas{
	\ell_{\text{det}} &\coloneqq \ell_{\text{det}}^{01}(\Ch_{\text{det}}, \Ch_\prp, \Ch_\prs, \Ch_\loc, x, y), \\
	\ell_\loc &\coloneqq \ell_\loc^{01}(\Ch_\loc, \Ch_\prp, x, y), \\
	\ell_\prs &\coloneqq \ell_\prs^{01}(\Ch_\prs, \Ch_\prp, x, y), \\
	E_\loc &\coloneqq \{ (x, y) \in \Xs\times \Ys \mid \ell_\loc = 1\}, \\
	E_\prs &\coloneqq \{ (x, y) \in \Xs\times \Ys \mid \ell_\prs = 1\}, \\
	U_{\text{det}} &\coloneqq \Xs\times \Ys \setminus E_\loc \setminus E_\prs.
}

\eqas{
L_{D}(\Ch_{\text{det}}; \Ch_\prp, \Ch_\prs, \Ch_\loc) 
&= \int \ell_{\text{det}} p(x, y) ~\mathrm{d}x \mathrm{d}y \\
&\le \int_{E_\loc} \ell_{\text{det}} p(x, y) ~\mathrm{d}x \mathrm{d}y + \int_{E_\prs} \ell_{\text{det}} p(x, y) ~\mathrm{d}x \mathrm{d}y + \int_{U_{\text{det}}} \ell_{\text{det}} p(x, y) ~\mathrm{d}x \mathrm{d}y \\
&= \int_{E_\loc} \ell_{\text{det}} p(x, y) ~\mathrm{d}x \mathrm{d}y + \int_{E_\prs} \ell_{\text{det}} p(x, y) ~\mathrm{d}x \mathrm{d}y \\
&\le \int_{E_\loc} p(x, y) ~\mathrm{d}x \mathrm{d}y + \int_{E_\prs} p(x, y) ~\mathrm{d}x \mathrm{d}y  \\
&= \int_{E_\loc} \ell_\loc p(x, y) ~\mathrm{d}x \mathrm{d}y + \int_{E_\prs} \ell_\prs p(x, y) ~\mathrm{d}x \mathrm{d}y \\
&= \int \ell_\loc p(x, y) ~\mathrm{d}x \mathrm{d}y + \int \ell_\prs p(x, y) ~\mathrm{d}x \mathrm{d}y \\
&= L_D(\Ch_\loc; \Ch_\prp) + L_D(\Ch_\prs; \Ch_\prp) \\
&\le \epsilon_\loc' + \epsilon_\prs',
}
where 
the second equality holds since $\Ch_{\text{det}}$ does not make error on $U_\text{det}$,
the last inequality holds with probability at least $1 - (\delta_\loc' + \delta_\prs')$ due to the union bound.

\section{Proof of Theorem \ref{thm:edge_composed}}
\label{apdx:thm:det_edge_proof}

Let 
\eqas{
    \ell_{\dtr, t} &\coloneqq \ell_\dtr^{01}(\Ch_\dtr, x_t, y_t) \\
    \ell_{\dtr, t+1} &\coloneqq \ell_\dtr^{01}(\Ch_\dtr, x_{t+1}, y_{t+1}) \\
    \ell_\edge &\coloneqq \ell_\edge^{01}(\Ch_\edge, \Ch_\dtr, x_t,x_{t+1}, y_t, y_{t+1} ) \\
    \ell_\edge^* &\coloneqq \ell_\edge^{01}(\Ch_\edge, C^*_\dtr, x_t,x_{t+1}, y_t, y_{t+1} ) \\
    E_{\dtr, t} &\coloneqq \{ (x_t, y_t) \in \Xs \times \Ys \mid \ell_{\dtr, t} = 1 \}, \\
    E_{\dtr, t+1} &\coloneqq \{ (x_{t+1}, y_{t+1}) \in \Xs \times \Ys \mid \ell_{\dtr, t+1} = 1\}, \\
    U_{\edge} &\coloneqq \Xs\times \Xs\times \Ys \times \Ys \setminus E_{\dtr, t} \times E_{\dtr, t+1}.
}
Then, the following holds:
\eqas{
&L_D(\Ch_\edge; \Ch_\dtr) \\
&= \Exp_{x_t, x_{t+1}, y_t, y_{t+1}}\[ \ell_{\edge}^{01}(\Ch_{\edge}, \Ch_{\det}, x_t, x_{t+1}, y_t, y_{t+1}) \] \\
&= \int \ell_\edge p(x_t, x_{t+1}, y_t, y_{t+1}) ~\mathrm{d} x_t \mathrm{d}x_{t+1} \mathrm{d}y_t \mathrm{d}y_{t+1} \\
&= \int_{E_{\dtr, t} \times E_{\dtr, t+1}} \ell_\edge p(x_t, x_{t+1}, y_t, y_{t+1}) ~\mathrm{d} x_t \mathrm{d}x_{t+1} \mathrm{d}y_t \mathrm{d}y_{t+1} + \int_{U_\edge} \ell_\edge p(x_t, x_{t+1}, y_t, y_{t+1}) ~\mathrm{d} x_t \mathrm{d}x_{t+1} \mathrm{d}y_t \mathrm{d}y_{t+1}.
}
Here, the first term is bounded as follows:
\eqas{
&\int_{E_{\dtr, t} \times E_{\dtr, t+1}} \ell_\edge p(x_t, x_{t+1}, y_t, y_{t+1}) ~\mathrm{d} x_t \mathrm{d}x_{t+1} \mathrm{d}y_t \mathrm{d}y_{t+1} \\
&\le \int_{E_{\dtr, t} \times E_{\dtr, t+1}} p(x_t, x_{t+1}, y_t, y_{t+1}) ~\mathrm{d} x_t \mathrm{d}x_{t+1} \mathrm{d}y_t \mathrm{d}y_{t+1} \\
&= \int_{E_{\dtr, t} \times E_{\dtr, t+1}} \frac{\ell_{\dtr, t} + \ell_{\dtr, t+1}}{2} p(x_t, x_{t+1}, y_t, y_{t+1}) ~\mathrm{d} x_t \mathrm{d}x_{t+1} \mathrm{d}y_t \mathrm{d}y_{t+1} \\
&= \frac{1}{2} \int_{E_{\dtr, t} \times E_{\dtr, t+1}} \ell_{\dtr, t} p(x_t, x_{t+1}, y_t, y_{t+1}) ~\mathrm{d} x_t \mathrm{d}x_{t+1} \mathrm{d}y_t \mathrm{d}y_{t+1} \\ &+ \frac{1}{2}\int_{E_{\dtr, t} \times E_{\dtr, t+1}} \ell_{\dtr, t+1} p(x_t, x_{t+1}, y_t, y_{t+1}) ~\mathrm{d} x_t \mathrm{d}x_{t+1} \mathrm{d}y_t \mathrm{d}y_{t+1} \\
&\le \frac{1}{2} \int_{E_{\dtr, t}} \ell_{\dtr, t} p(x_t, y_t) ~\mathrm{d} x_t \mathrm{d}y_t + \frac{1}{2}\int_{E_{\dtr, t+1}} \ell_{\dtr, t+1} p(x_{t+1}, y_{t+1}) ~\mathrm{d}x_{t+1} \mathrm{d}y_{t+1} \\
&= \frac{1}{2} \int \ell_{\dtr, t} p(x_t, y_t) ~\mathrm{d} x_t \mathrm{d}y_t + \frac{1}{2} \int \ell_{\dtr, t+1} p(x_{t+1}, y_{t+1}) ~\mathrm{d}x_{t+1} \mathrm{d}y_{t+1} \\
&= \frac{1}{2} L_D(\Ch_\dtr) + \frac{1}{2} L_D(\Ch_\dtr) \\
&\le \epsilon_\dtr,
}
where the last inequality holds with probability at least $1 - \delta_\dtr$.
The second term is bounded as follows:
\eqas{
\int_{U_\edge} \ell_\edge p(x_t, x_{t+1}, y_t, y_{t+1}) ~\mathrm{d} x_t, \mathrm{d}x_{t+1}, \mathrm{d}y_t, \mathrm{d}y_{t+1} 
&= \int_{U_\edge} \ell_\edge^* p(x_t, x_{t+1}, y_t, y_{t+1}) ~\mathrm{d} x_t, \mathrm{d}x_{t+1}, \mathrm{d}y_t, \mathrm{d}y_{t+1} \\
&\le \int \ell_\edge^* p(x_t, x_{t+1}, y_t, y_{t+1}) ~\mathrm{d} x_t, \mathrm{d}x_{t+1}, \mathrm{d}y_t, \mathrm{d}y_{t+1} \\
&= L_D(\Ch_\edge; C^*_\dtr) \\
&\le \epsilon_\edge,
}
where the last inequality holds with probability at least $1 - \delta_\edge$.
Thus, the claim follows.

\section{Auxiliary Detection and Tracking Results}
\label{apdx:emp_aux}
Please see Figure \ref{fig:ps_comp_aux}.

\begin{figure*}[hbt!]
\centering
\begin{subfigure}[b]{0.32\linewidth}
\centering
\includegraphics[width=\linewidth]{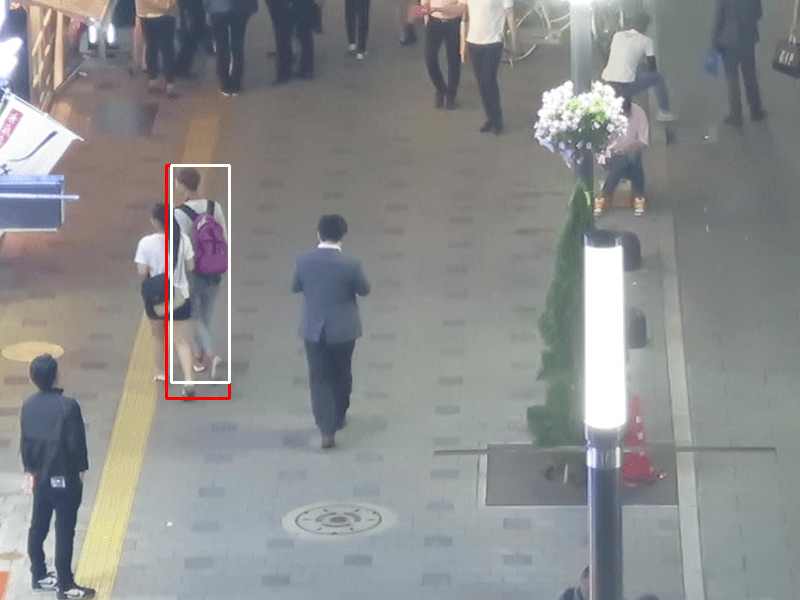} 
\end{subfigure}
\begin{subfigure}[b]{0.32\linewidth}
\centering
\includegraphics[width=\linewidth]{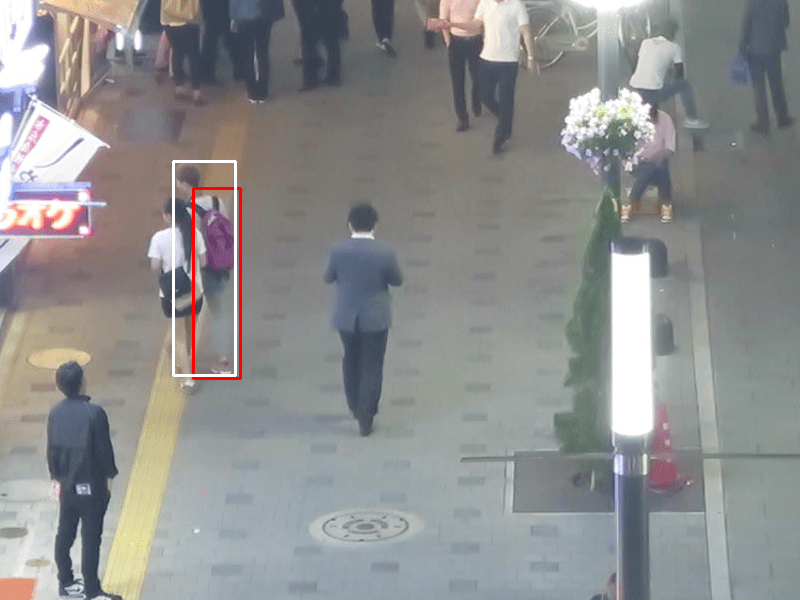} 
\end{subfigure}
\begin{subfigure}[b]{0.32\linewidth}
\centering
\includegraphics[width=\linewidth]{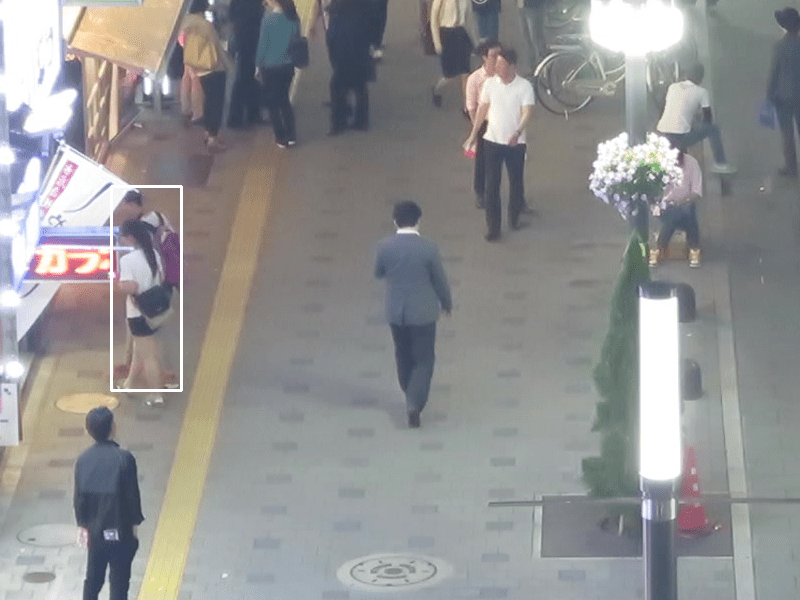} 
\end{subfigure}

\begin{subfigure}[b]{0.32\linewidth}
\centering
\includegraphics[width=\linewidth]{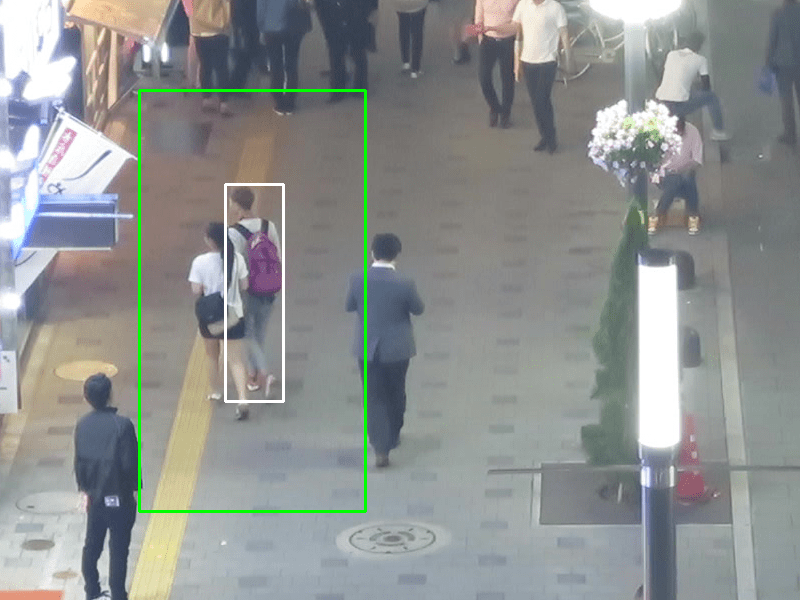} 
\end{subfigure}
\begin{subfigure}[b]{0.32\linewidth}
\centering
\includegraphics[width=\linewidth]{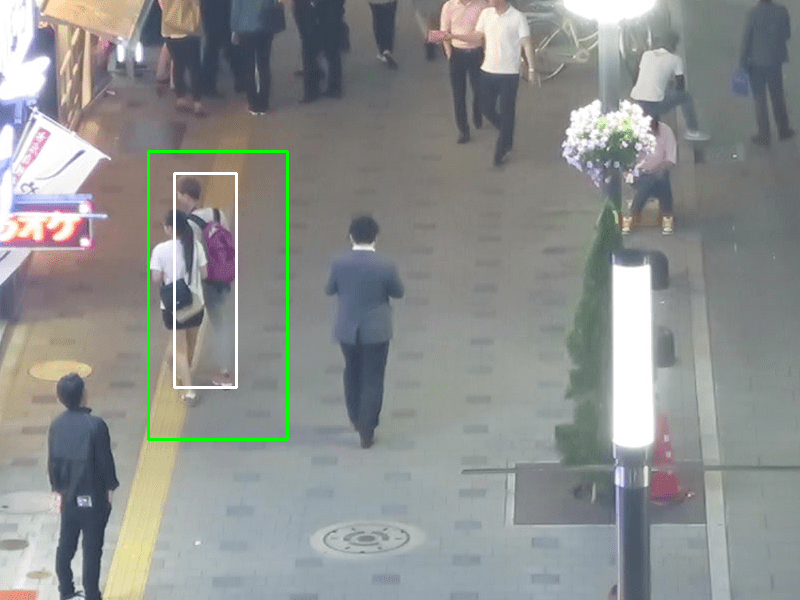} 
\end{subfigure}
\begin{subfigure}[b]{0.32\linewidth}
\centering
\includegraphics[width=\linewidth]{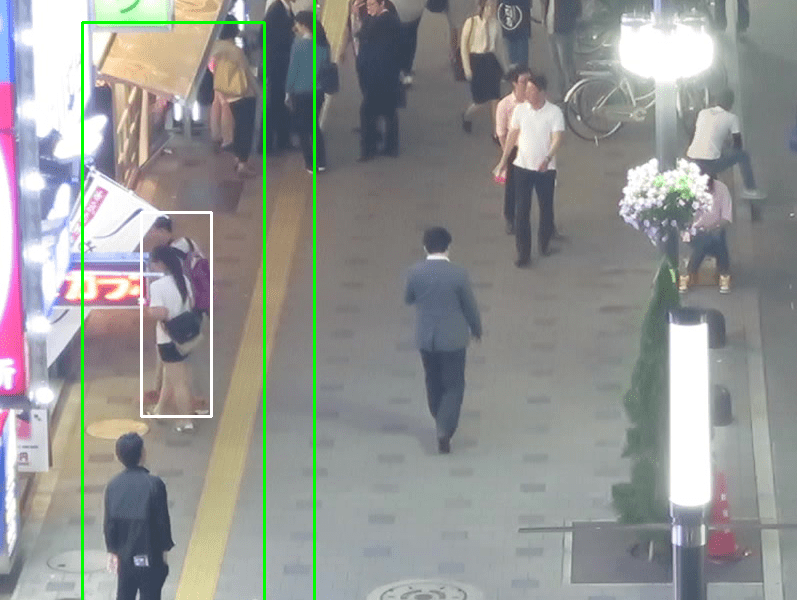} 
\end{subfigure}

\begin{subfigure}[b]{0.32\linewidth}
\centering
\includegraphics[width=\linewidth]{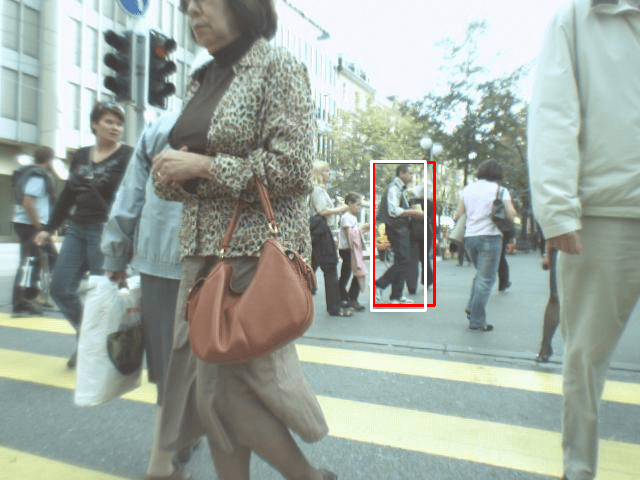} 
\end{subfigure}
\begin{subfigure}[b]{0.32\linewidth}
\centering
\includegraphics[width=\linewidth]{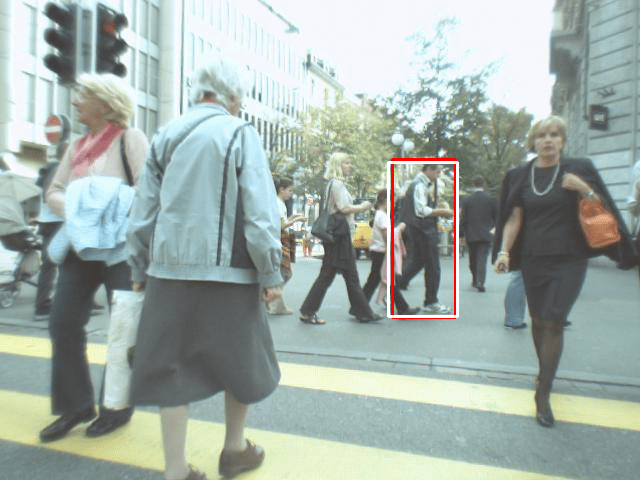} 
\end{subfigure}
\begin{subfigure}[b]{0.32\linewidth}
\centering
\includegraphics[width=\linewidth]{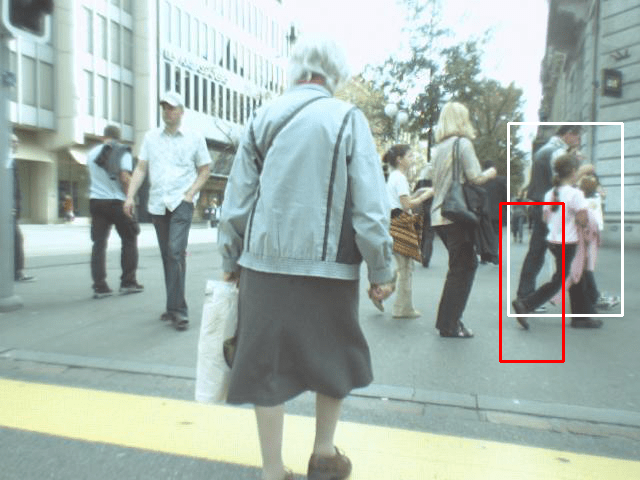} 
\end{subfigure}

\begin{subfigure}[b]{0.32\linewidth}
\centering
\includegraphics[width=\linewidth]{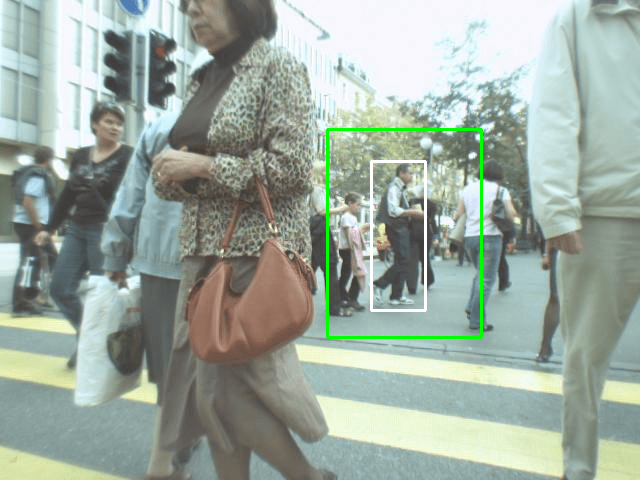} 
\end{subfigure}
\begin{subfigure}[b]{0.32\linewidth}
\centering
\includegraphics[width=\linewidth]{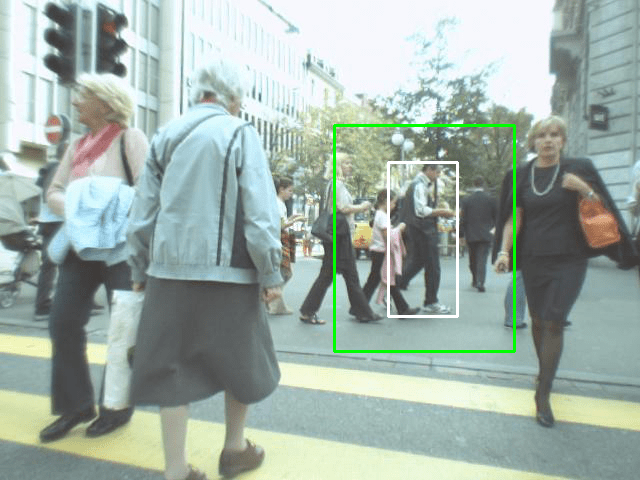} 
\end{subfigure}
\begin{subfigure}[b]{0.32\linewidth}
\centering
\includegraphics[width=\linewidth]{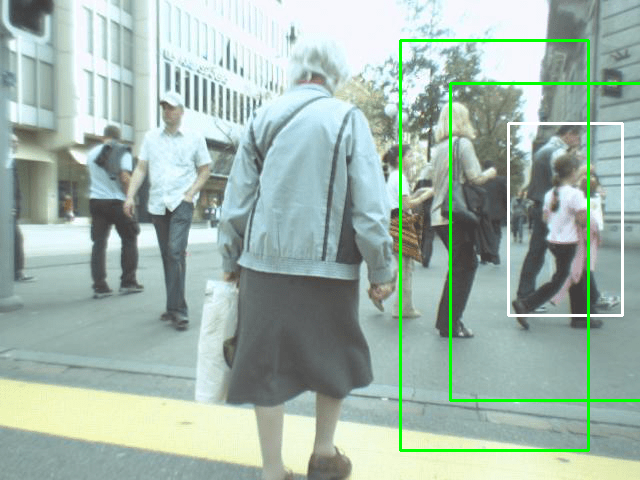} 
\end{subfigure}



\label{fig:ps_comp_aux}
\caption{Composition of detection prediction set and edge prediction sets on \textit{Sequence-04, 
Sequence-05}. White boxes are ground truth, red boxes are bounding boxes of tracks in Tracktor's detection, and green boxes are our edge prediction sets $C_{\hat\tau_\edge}$. Row 1 are detection and tracking results from Tracktor, and row 2 are results from our method. In these cases, our detection prediction sets contain the ground truth object bounding boxes with high probability, and our edge prediction sets contain the track that tracks our interested object with high probability, but Tracktor does not.}
\label{fig:ps_comp_aux}
\end{figure*}

\section{Experiment Setup}
We integrate our detection and edge prediction sets with Tracktor's heuristics, i.e., using previously detected bounding boxes as region proposals, utilizing Siamese features to accomplish re-identification. Specifically, we first augment the vanilla Faster RCNN in Tracktor with our detection prediction set. The modified Faster RCNN will return bounding boxes whose proposals are in the presence prediction set $C_{\hat\tau_\prs}(\cdot)$ and the location prediction set $C_{\hat\tau_\loc}(\cdot)$. We then compute edge prediction set $\Ch_\edge(\cdot)$ for each detected object by computing intersection over union (IoU) between detected bounding boxes at adjacent time steps. In this way, we use the detection prediction set to estimate bounding boxes for each object; we leverage the edge prediction set to track objects. 

Besides integrating our prediction sets into the Tracktor implementation, we also modify some of Tracktor's heuristics. In particular, we do Non-Maximum Suppression (NMS) to bounding boxes from Tracktor's heuristics detection instead of Faster RCNN's detection, which is in the reverse order in Tracktor. Also, we set more conservative thresholds for Tracktor's heuristics, i.e., 0.1 for person detection, 0.6 for non-maximum detection between bounding boxes in Tracktor's heuristics detection and detection prediction sets. Please find our implementation from this link: \url{https://drive.google.com/file/d/1-nn3LywrtyqKZTkc5Rs3J1wYkFb6lmMX/view?usp=sharing}.



\end{document}